\documentclass[11pt]{article}

\usepackage[ansinew]{inputenc}
\usepackage{amsmath,amssymb}
\usepackage{color}
\usepackage{a4wide}
\usepackage{verbatim}
\usepackage{scrtime}
\usepackage{float}
\usepackage{graphicx}
\usepackage{graphics}
\usepackage{pstricks}
\usepackage{fancyhdr}
\usepackage{subfigure}
\usepackage{amsthm}
\usepackage{url}

\large\normalsize

\newcommand{\change}[1]{#1}

\newcommand{\trace}{{\rm Tr}}

\newcommand{\SVD}{{\rm SVD}}
\newcommand{\rank}{{\rm rank}}

\newcommand{\Sym}{{\mathrm{Sym}}}
\newcommand{\Skew}{{\mathrm{Skew}}}

\newcommand{\rc}{\nabla}
\newcommand{\D}{\mathrm{D}}

\newcommand{\Grassmann}[2]{{\mathrm{Gr}({#1},{#2})}}
\newcommand{\ConePD}[1]{S_{++}({#1})}

\newcommand{\FixedRank}[3]{\mathbb{R}_{#1}^{{#2}\times{#3}}}

\newcommand{\OG}[1]{{\mathcal{O}({#1})}}
\newcommand{\GL}[1]{{\mathrm{GL}({#1})}}
\newcommand{\Stiefel}[2]{{\mathrm{St}({#1},{#2})}}

\newcommand{\set}[2]{{\{{#1}:\ {#2}\}}}

\newcommand{\argmin}{\operatornamewithlimits{arg\,min}}

\newcommand{\grad}{\mathrm{grad}}
\newcommand{\hess}{\mathrm{Hess}}

\newcommand{\Grad}{\mathrm{Grad}}

\newcommand{\mat}[1]{{\bf #1}}
\renewcommand{\vec}[1]{{\bf #1}}

\newcommand*\samethanks[1][\value{footnote}]{\footnotemark[#1]}

\newcommand{\subject}{\mathrm{subject\  to\quad}}

\newcommand{\symmetric}[1]{S_{sym}({#1})}

\title{Fixed-rank matrix factorizations and\\ Riemannian low-rank optimization\thanks{This paper presents research results of the Belgian Network DYSCO (Dynamical Systems, Control, and Optimization), funded by the Interuniversity Attraction Poles Programme, initiated by the Belgian State, Science Policy Office. The scientific responsibility rests with its authors. Bamdev Mishra is a research fellow of the Belgian National Fund for Scientific Research (FNRS).}
}
\author{B.~Mishra\thanks{Department of Electrical Engineering and Computer Science, University of Li\`ege, 4000 Li\`ege,
Belgium (B.Mishra@ulg.ac.be, Gillesmy@gmail.com, R.Sepulchre@ulg.ac.be).}
        \and G.~Meyer\samethanks[2]
        \and S.~Bonnabel\thanks{Robotics center Mines ParisTech Boulevard Saint-Michel, 60, 75272 Paris, France (Silvere.Bonnabel@mines-paristech.fr).}
        \and R.~Sepulchre\samethanks[2]  \thanks{ORCHESTRON, INRIA-Lille, Lille, France}
        }

\begin{document}

\maketitle

\begin{abstract}
Motivated by the problem of learning a linear regression model whose parameter is a large fixed-rank non-symmetric matrix, we consider the optimization of a smooth cost function defined on the set of fixed-rank matrices. We adopt the geometric framework of optimization on Riemannian quotient manifolds. We study the underlying geometries of several well-known fixed-rank matrix factorizations and then exploit the Riemannian quotient geometry of the search space in the design of a class of gradient descent and trust-region algorithms. The proposed algorithms generalize our previous results on fixed-rank symmetric positive semidefinite matrices, apply to a broad range of applications, scale to high-dimensional problems and confer a geometric basis to recent contributions on the learning of fixed-rank non-symmetric matrices. We make connections with existing algorithms in the context of low-rank matrix completion and discuss relative usefulness of the proposed framework. Numerical experiments suggest that the proposed algorithms compete with the state-of-the-art and that manifold optimization offers an effective and versatile framework for the design of machine learning algorithms that learn a fixed-rank matrix.

\end{abstract}

\section{Introduction}
The problem of learning a low-rank matrix is a fundamental problem arising in many modern machine learning applications such as collaborative filtering \cite{rennie05a}, classification with multiple classes \cite{amit07a}, learning on pairs \cite{abernethy09a}, dimensionality reduction \cite{cai07a}, learning of low-rank distances \cite{kulis09a, meyer11c} and low-rank similarity measures \cite{shalit10a}, multi-task learning \cite{evgeniou05b, mishra11a}, to name a few. Parallel to the development of these new applications, the ever-growing size and number of large-scale datasets demands machine learning algorithms that can cope with large matrices. Scalability to high-dimensional problems is therefore a crucial issue in the design of algorithms that learn a low-rank matrix. Motivated by the above applications, the paper focuses on the following optimization problem
\begin{equation}\label{eq:intro-general-formulation}
	\min_{\mat{W} \in\FixedRank{r}{d_1}{d_2}} f(\mat{W}),
\end{equation}
where $f: \mathbb{R}^{d_1 \times d_2} \rightarrow \mathbb{R}$ is a smooth cost function and the search space is the set of fixed-rank non-symmetric real matrices,
\begin{equation*}
	\FixedRank{r}{d_1}{d_2} = \{\mat{W}\in\mathbb{R}^{d_1\times d_2} : \rank(\mat{W}) = r\}.
\end{equation*}
\change{A particular case of interest is when $r\ll \min(d_1, d_2)$.} In Section \ref{sec:applications} we show that the considered optimization problem \eqref{eq:intro-general-formulation} encompasses various modern machine learning applications. We tackle problem (\ref{eq:intro-general-formulation}) in a Riemannian framework, that is, by solving an unconstrained optimization on a Riemannian manifold in bijection with the nonlinear space $\FixedRank{r}{d_1}{d_2}$. \change{This nonlinear space is an abstract space that is given the structure of a Riemannian quotient manifold in Section \ref{sec:quotient_geometry}. The search space is motivated as a product space of well-studied manifolds which allows to derive the geometric notions in a straightforward and systematic way. Simultaneously, it ensures that we have \emph{enough} flexibility in combining the different \emph{pieces} together. One such flexibility is the choice of metric on the product space.}

The paper follows and builds upon a number of recent contributions in that direction: the Ph.D. thesis \cite{meyer11b} and several papers by the authors \cite{meyer11a, meyer11c, mishra11a, mishra11b, mishra12a, journee09a}. The main contribution of this paper is to emphasize the common framework that underlines those contributions, with the aim of illustrating the versatile framework of Riemannian optimization for \change{rank-constrained} optimization. Necessary ingredients to perform both first-order and second-order optimization are listed \change{for ready referencing}. \change{We discuss three popular fixed-rank matrix factorizations that embed the rank constraint. Two of these factorizations have been studied individually in \cite{meyer11a, mishra11b}. Exploiting the third factorization (the subspace-projection factorization in Section \ref{sec:factorizations-subspace-proj}) in the Riemannian framework is new.} An attempt is also made to classify the existing algorithms into various geometries and show the common structure that connects them all. Scalability \change{of both first-order and second-order optimization algorithms} to large dimensional problems is shown in Section \ref{sec:numerical_comparisons}. 

The paper is organized as follows. Section \ref{sec:applications}  provides some concrete motivation for the proposed fixed-rank optimization problem.  Section \ref{sec:quotient_spaces}
reviews three classical fixed-rank matrix factorizations and introduces the quotient nature of the  underlying search spaces. Section  \ref{sec:quotient_geometry} develops the Riemannian quotient geometry of these three search spaces, providing all the concrete matrix operations required to code any first-order or second-order algorithm. Two basic algorithms are further detailed in Section \ref{sec:algorithms}. They underlie all numerical tests presented in Section \ref{sec:numerical_comparisons}.


\section{Motivation and applications}\label{sec:applications}
In this section, a number of modern machine learning applications are cast as an optimization problem on the set of fixed-rank non-symmetric matrices. 

\subsection{Low-rank matrix completion}\label{sec:applic-completion}
The problem of low-rank matrix completion amounts to estimating the missing entries of a matrix from a limited number of its entries. There has been a large number of research contributions on this subject over the last few years, addressing the problem both from a theoretical \cite{candes08b, gross11a} and from an algorithmic point of view \cite{rennie05a, cai10a, lee09a, meka09a, keshavan10a, simonsson10a, jain10a, mazumder10a, boumal11a, ngo12a}. An important and popular application of the low-rank matrix completion problem is collaborative filtering \cite{rennie05a, abernethy09a}.

Let $\mat{W}^{\star}\in\mathbb{R}^{d_1\times d_2}$ be a matrix whose entries $\mat{W}^{\star}_{ij}$ are only given for some indices $(i,j)\in\Omega$, where $\Omega$ is a subset of the complete set of indices $\{(i,j):i\in\{1,\dots,d_1\}\text{ and }j\in\{1,\dots,d_2\}\}$. \change{Fixed-rank} matrix completion amounts to solving the following optimization problem
\begin{equation}\label{eq:matrix-completion-formulation}
\begin{array}{llll}
	\min\limits_{\mat{W}\in\mathbb{R}^{d_1\times d_2}}
		&	\frac{1}{|\Omega|}\|\mathcal{P}_{\Omega}(\mat{W}) - \mathcal{P}_{\Omega}(\mat{W}^{\star})\|_F^2 \\
	 \subject & \rank(\mat{W})=r,
\end{array}
\end{equation}
where the function \change{$\mathcal{P}_{\Omega}(\mat{W})_{ij}=\mat{W}_{ij}$} if $(i,j) \in \Omega$ and \change{$\mathcal{P}_{\Omega}(\mat{W})_{ij}=0$} otherwise and the norm $\|\cdot \|_F$ is \emph{Frobenius} norm. $\mathcal{P}_{\Omega}$ is also called the \emph{orthogonal sampling operator} and $|\Omega|$ is the cardinality of the set $\Omega$ (equal to the number of known entries).

The rank constraint captures redundant patterns in $\mat{W}^{\star}$ and ties the known and unknown entries together. The number of given entries $|\Omega|$ is \change{of $O(d_1r + d_2r - r^2)$ which is} much smaller than $d_1d_2$ \change{(the total number of entries in $\mat{W}^*$) when $r \ll \min(d_1, d_2)$}. Recent contributions provide conditions on $|\Omega|$ under which exact reconstruction is possible from entries sampled uniformly and at random \cite{candes08b, keshavan10a}. \change{An application of this is in movie recommendations. The matrix to complete is a matrix of movie ratings of different users; a very sparse matrix with few ratings per user. The predictions of unknown ratings with a low-rank prior would have the interpretation that users' preferences only depend on few \emph{genres} \cite{netflix06a}.}

\subsection{Learning on data pairs}
\change{The problem of learning on data pairs amounts to learning a predictive model $\hat{y}:\mathcal{X}\times \mathcal{Z}\rightarrow\mathbb{R}$ from $n$ training examples $\{(\vec{x}_i,\vec{z}_i,y_i)\}_{i=1}^{n}$ where data $\vec{x}_i$ and $\vec{z}_i$ are associated with two types of samples drawn from the set $\mathcal{X} \times \mathcal{Z}$ and ${y}_i \in \mathbb{R}$ is the associated scalar observation from the predictive model. If the predictive model is the bilinear form $\hat{y}=\vec{x}^T\mat{W}\vec{z}$ with $\mat{W}\in\FixedRank{r}{d_1}{d_2}$, $\vec{x} \in \mathbb{R}^{d_1}$ and $\vec{z} \in \mathbb{R}^{d_2}$, then the problem boils down to the optimization problem,}
\begin{equation}
	\min_{\mat{W}\in\FixedRank{r}{d_1}{d_2}} \quad \frac{1}{n}\sum_{i = 1}^n \ell(\vec{x}_i^T \mat{W} \vec{z}_i,y_i),
\end{equation}
where the loss function $\ell$ penalizes the discrepancy between a scalar (experimental) observation $y$ and the predicted value $\hat{y}$.

An application of this setup is the inference of edges in bipartite or directed graphs. Such problems arise in bioinformatics for the identification of interactions between drugs and target proteins, micro-RNA and genes or genes and diseases \cite{yamanishi08a,bleakley09a}. Another application is concerned with image domain adaptation \cite{kulis11a} where a transformation $\vec{x}^T\mat{W}\vec{z}$ is learned between labeled images $\vec{x}$ from a source domain $\mathcal{X}$ and labeled images $\vec{z}$ from a target domain $\mathcal{Z}$. The transformation $\mat{W}$ maps new input data from one domain to the other. A potential interest of the rank constraint in these applications is to address problems with a high-dimensional feature space and perform dimensionality reduction on the two data domains.

\subsection{Multivariate linear regression}
\change{In multivariate linear regression}, given matrices $\mat{Y}\in \mathbb{R}^{n \times k}$ (output space) and $\mat{X} \in \mathbb{R}^{n\times q}$ (input space), we seek to learn a weight/coefficient matrix $\mat{W} \in\FixedRank{r}{q}{k}$ that minimizes the discrepancy between $\mat{Y}$ and $\mat{XW}$ \cite{yuan07a}. Here $n$ is the number of observations, $q$ is the number of predictors and $k$ is the number of responses. 

One popular approach to multivariate linear regression problem is by minimizing a \emph{quadratic loss} function. Note that in various applications \emph{responses} are related and may therefore, be represented with much fewer coefficients \change{\cite{yuan07a, amit07a}}. \change{This corresponds to finding the best low-rank matrix such that}
\[
\min_{\mat{W}\in\FixedRank{r}{q}{k}} \quad \| \mat{Y} - \mat{XW}\|_F^2 .
\]
Though the quadratic loss function is shown here, the optimization setup extends to other smooth loss functions as well.

\change{An application of this setup  in financial econometrics is considered in \cite{yuan07a} where the future returns of assets are estimated on the basis of their historical performance using the above formulation.}


\section{Matrix factorization and quotient spaces}\label{sec:quotient_spaces}
\change{A popular way to parameterize fixed-rank matrices   is through matrix factorization. We review three popular matrix factorizations for fixed-rank non-symmetric matrices and study the underlying Riemannian geometries of the resulting search space. }

\begin{figure}[htbp]
	\centering
	\includegraphics[scale = 0.65]{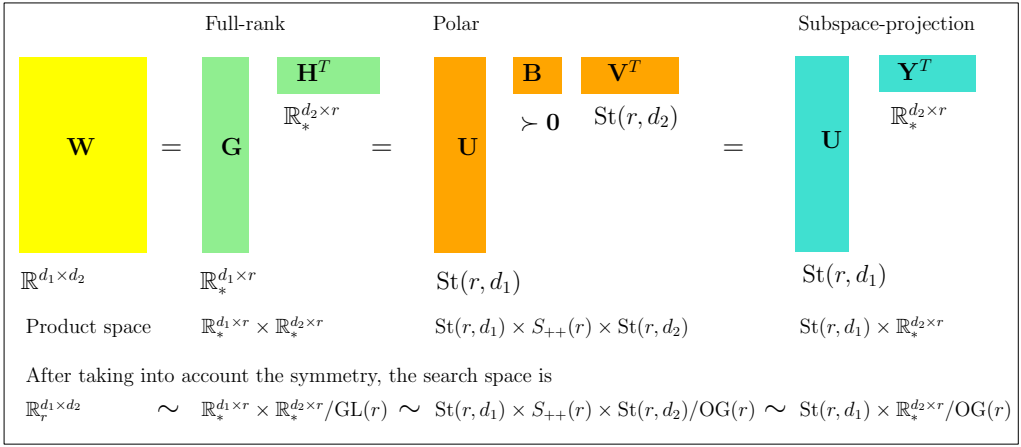}
	\caption{Fixed-rank matrix factorizations lead to quotient search spaces due to intrinsic \emph{symmetries}. The pictures emphasize the situation of interest, i.e., the rank $r$ is small compared to the matrix dimensions.}
	\label{fig:factorizations}
\end{figure}
The three fixed-rank matrix factorizations of interest all arise from the thin singular value decomposition of a rank-$r$ matrix $\mat{W} = \mat{U}\mat{\Sigma}\mat{V}^T$, where $\mat{U}$ is a $d_1 \times r$ matrix with orthogonal columns, that is, an element of the Stiefel manifold $\Stiefel{r}{d_1}=\set{\mat{U}\in\mathbb{R}^{d_1\times r}}{\mat{U}^T\mat{U} = \mat{I}}$, $\mat{\Sigma} \in {\rm Diag}_{++}(r)$ is a $r\times r$ diagonal matrix with positive entries and $\mat{V}\in\Stiefel{r}{d_2}$. The singular value decomposition (SVD) exists for any matrix $\mat{W}\in\FixedRank{r}{d_1}{d_2}$ \cite{golub06a}.

\subsection{Full-rank factorization (beyond Cholesky-type decomposition)}\label{sec:factorizations-balanced}
The most popular low-rank factorization is obtained when the singular value decomposition (SVD) is rearranged as
\begin{equation*}
	\mat{W} = (\mat{U}\mat{\Sigma}^{\frac{1}{2}})(\mat{\Sigma}^{\frac{1}{2}}\mat{V}^T) = \mat{G}\mat{H}^T,
\end{equation*}
where $\mat{G}= \mat{U}\mat{\Sigma}^{\frac{1}{2}} \in\mathbb{R}_*^{d_1\times r}$, $\mat{H}=\mat{V}\mat{\Sigma}^{\frac{1}{2}}\in\mathbb{R}_*^{d_2\times r}$ and $\mathbb{R}^{d\times r}_{*}$ is the set of full column rank $d \times r$ matrices, also known as \emph{full-rank matrix factorization}. The resulting factorization is not unique because the transformation,
\begin{equation}\label{eq:balanced-group-action}
	(\mat{G},\mat{H})\mapsto (\mat{G}\mat{M}^{-1},\mat{H}\mat{M}^{T}),
\end{equation}
where $\mat{M}\in\GL{r}=\set{\mat{M}\in\mathbb{R}^{r\times r}}{\det(\mat{M})\neq 0}$, leaves the original matrix $\mat{W}$ unchanged \cite{piziak99a}. \change{This symmetry comes from the fact that the row and column spaces are invariant to the change of coordinates}. The classical remedy to remove this indeterminacy \change{in the case of symmetric positive semidefinite matrices} is the Cholesky factorization, which imposes further (triangular-like) structure in the factors. \change{The LU decomposition plays a similar role for the non-symmetric} matrices \cite{golub06a}. In a manifold setting, we instead encode the invariance map \eqref{eq:balanced-group-action} in an abstract search space by optimizing over a set of equivalence classes defined as 
\begin{equation}\label{eq:equivalence-classes-balanced}
 [\mat{W}] = [(\mat{G},\mat{H})] = \set{(\mat{G}\mat{M}^{-1},\mat{H}\mat{M}^{T})}{\mat{M}\in\mathrm{GL}(r)},
\end{equation}
instead of the product space $\mathbb{R}_*^{d_1\times r} \times \mathbb{R}_*^{d_2\times r}$. The set of equivalence classes  is denoted as 
\begin{equation}\label{eq:quotient-balanced}
	\mathcal{W}:= \overline{\mathcal{W}}/\GL{r}.
\end{equation}

\change{The product space $\mathbb{R}_*^{d_1 \times r} \times \mathbb{R}_*^{d_2 \times r}$ is called the \emph{total space}, denoted by $\overline{\mathcal W}$. The set ${\rm GL} (r)$ is called the \emph{fiber space}. The set of equivalence classes $\mathcal W$ is called the quotient space. In the next section it is given the structure of a Riemannian manifold over which optimization algorithms are developed.}

\subsection{Polar factorization (beyond SVD)}\label{sec:factorizations-polar}
The second quotient structure for the set $\FixedRank{r}{d_1}{d_2}$ is obtained by considering the following group action on the SVD \cite{bonnabel09a},
\begin{equation*}
	(\mat{U},\mat{\Sigma},\mat{V}) \mapsto (\mat{U}\mat{O},\mat{O}^T\mat{\Sigma}\mat{O},\mat{V}\mat{O}),
\end{equation*}
where $\mat{O}$ is any $r \times r$ orthogonal matrix, that is, any element of the set
\begin{equation*}
	\OG{r} = \{\mat{O}\in\mathbb{R}^{r\times r}: \mat{O}^T\mat{O}=\mat{O}\mat{O}^T=\mat{I}\}.
\end{equation*}
This results in \emph{polar factorization}
\begin{equation*}
	\mat{W} = \mat{U}\mat{B}\mat{V}^T,
\end{equation*}
where $\mat{B}$ is now a $r \times r$ symmetric positive definite matrix, that is, an element of
\begin{equation}
	\ConePD{r} = \{\mat{B}\in\mathbb{R}^{r\times r}: \mat{B}^T = \mat{B}\succ 0\}.
\end{equation}
\change{The polar factorization reflects the original geometric purpose of singular value decomposition as representing an arbitrary linear transformation as the composition of two isometries and a scaling \cite{golub06a}. Allowing the scaling $\mat{B}$ to be positive definite rather than diagonal gives more flexibility in the optimization and removes the discrete symmetries induced by interchanging the order on the singular values. Empirical evidence to support the choice of $\ConePD{r}$ over ${\rm Diag}_{++}(r)$ (set of diagonal matrices with positive entries) for the middle factor $\mat{B}$ is shown in Section \ref{sec:polar_vs_svd}. The resulting search space is again the set of equivalence classes defined by}
\begin{equation}\label{eq:equivalence-classes-polar}
	[\mat{W}] = [(\mat{U},\mat{B},\mat{V})] 
		= \set{(\mat{U}\mat{O},\mat{O}^T\mat{B}\mat{O},\mat{V}\mat{O})}{\mat{O}\in\OG{r}}.
\end{equation}

\change{The total space is now $\overline{\mathcal W} = \Stiefel{r}{d_1} \times \ConePD{r} \times \Stiefel{r}{d_2}$. The fiber space is $\OG{r}$ and the resulting quotient space is, thus, the set of equivalence classes }
\begin{equation}\label{eq:quotient-polar}
\mathcal{W} = \overline{\mathcal W} / \OG{r}.
\end{equation}

\subsection{Subspace-projection factorization (beyond QR decomposition)}\label{sec:factorizations-subspace-proj}
The third low-rank factorization is obtained from the SVD when two factors are grouped together,
\begin{equation*}
	\mat{W} = \mat{U}(\mat{\Sigma}\mat{V}^T) = \mat{U}\mat{Y}^T,
\end{equation*}
where $\mat{U}\in\Stiefel{r}{d_1}$ and $\mat{Y}\in\mathbb{R}_*^{d_2 \times r}$ and is referred to as \emph{subspace-projection} factorization. The column subspace of $\mat{W}$ matrix is represented by $\mat{U}$ while $\mat{Y}$ is the (left) \emph{projection} or \emph{coefficient} matrix of $\mat{W}$. The factorization is not unique as it is invariant with respect to the group action $(\mat{U},\mat{Y})\mapsto(\mat{U}\mat{O},\mat{Y}\mat{O})$, whenever $\mat{O}\in\OG{r}$. The classical remedy to remove this indeterminacy is the QR factorization for which $\mat{Y}$ is chosen upper triangular \cite{golub06a}.  Here again we work with the set of equivalence classes
\begin{equation}\label{eq:equivalence-classes-subspace-proj}
	[\mat{W}] = [(\mat{U},\mat{Y})] = \set{(\mat{U}\mat{O},\mat{Y}\mat{O})}{\mat{O}\in\OG{r}}.
\end{equation}
\change{The search space is the quotient space}
\begin{equation}\label{eq:quotient-manifold-subspace-proj}
\mathcal{W} = \overline{\mathcal{W}} /\OG{r},
\end{equation}
\change{where the total space is $\overline{\mathcal{W}} := \Stiefel{r}{d_1} \times \mathbb{R}_* ^{d_2 \times r}$ and the fiber space is $\OG{r}$.} Recent contributions using this factorization include \cite{boumal11a, simonsson10a}.

\section{Fixed-rank matrix spaces as Riemannian submersions}\label{sec:quotient_geometry}
The general philosophy of optimization on manifolds is to recast a constrained  optimization problem in the Euclidean space $\mathbb{R}^{n}$ into an unconstrained optimization on a nonlinear search space that encodes the constraint. For special constraints that are sufficiently structured, the framework leads to an efficient computational framework \cite{absil08a}. The three total spaces considered in the previous section all admit product structures of well-studied differentiable manifolds $\Stiefel{r}{d_1}$, $\mathbb{R}_*^{d_1 \times r}$ and $\ConePD{r}$. Similarly, the fiber spaces are the Lie groups $\GL{r}$ and $\OG{r}$. In this section, all the quotient spaces of the three fixed-rank factorizations are shown to have the differential structure of a Riemannian quotient manifold.

Each point on a quotient manifold represents an entire equivalence class of matrices \change{in the total space}. Abstract geometric objects on the quotient manifold can be defined by means of matrix representatives. Below we show the development of various geometric objects that are are required to optimize a smooth cost function on the quotient manifold. Most of these notions follow directly from \cite[Chapters~3 and ~4]{absil08a}. In Table \ref{tab:matrix_representations} to \ref{tab:gradient_Hessian} we give the matrix representations of various geometric notions that are required to optimize a smooth cost function on a quotient manifold. More details of the matrix factorizations, full-rank factorization (Section \ref{sec:factorizations-balanced}) and polar factorization (Section \ref{sec:factorizations-polar}) may be found in \cite{mishra11a, meyer11b}. \change{The corresponding geometric notions for the subspace-projection factorization (Section \ref{sec:factorizations-subspace-proj}) are new to the paper but nevertheless, the development follows similar lines.}

\begin{figure}[t]
 	\centering
 	\begin{tabular}{cc}
 		\includegraphics[scale=.45]{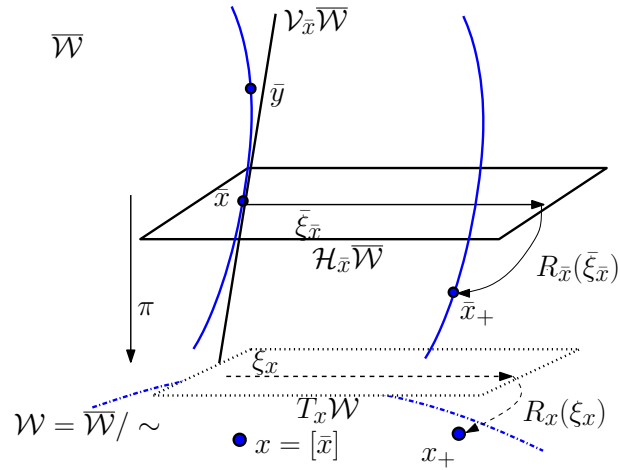}
 	\end{tabular}
 	\caption{Visualization of a Riemannian quotient manifold. The points $\bar{y}$ and $\bar{x}$ in the total space $\overline{ \mathcal{W}}$ belong to the same equivalence class and they represent a single point $[x]$ in the quotient space $\mathcal{W}$. $\pi : \overline{\mathcal{W}} \rightarrow \mathcal{W}$ is a Riemannian submersion. The subspaces $\mathcal{V}_{\bar x} \overline{\mathcal{W}}$ and $\mathcal{H}_{\bar x} \overline{\mathcal{W}}$ are complementary spaces of $T_{\bar x} \overline{\mathcal{W}}$. The horizontal space $\mathcal{H}_{\bar{x}}\overline{\mathcal{W}}$ provides a matrix representation to the abstract tangent space $T_{x} \mathcal{W}$ of the Riemannian quotient manifold. The mapping $R_{\bar x}$ maps a horizontal vector onto the total space.}
 	\label{fig:quotient_optimization}
 \end{figure}

\subsection{Quotient manifold representation}
Consider a total space $\overline{\mathcal{W}}$  equipped with an equivalence relation $\sim$. The equivalence class of a given point ${\bar x} \in \overline{\mathcal W}$ is the set $[\bar x] = \{{\bar y} \in \overline{\mathcal W}: \bar{y} \sim \bar{x}  \}$. The set $\mathcal W$ of all equivalence classes  is the quotient manifold of $\overline{\mathcal W}$ by the equivalence relation $\sim$.  The mapping $\pi: \overline{\mathcal W} \rightarrow \mathcal{W}$ is called the natural or canonical projection map. In Figure \ref{fig:quotient_optimization}, we have $\pi({\bar x}) = \pi({\bar y})$ if and only if $\bar{x } \sim {\bar y}$ and therefore, $[{\bar x}] = \pi^{-1}(\pi( {\bar x}))$. We represent an element of the quotient space $\mathcal{W}$ by $x = [\bar{x}]$ and its matrix representation in the total space $ \overline{\mathcal W}$ by ${\bar x}$.

In Section \ref{sec:quotient_spaces} we see that the total spaces for the three fixed-rank matrix factorizatons are in fact, different product spaces of the set of full column rank matrices $\mathbb{R}_*^{d_1 \times r}$,  the stet of matrices of size $d_1 \times r$ with orthonormal columns $\Stiefel{r}{d_1}$ \cite{edelman98a}, and the set of positive definite $r\times r$ matrices $\ConePD{r}$ \cite{bhatia07a}. Each of these manifolds is a smooth homogeneous space and their product structure preserves the smooth differentiability property \cite[Section~3.1.6]{absil08a}. 

The quotient spaces of the three matrix factorizations are given by the equivalence relationships shown in (\ref{eq:quotient-balanced}), (\ref{eq:quotient-polar}) and (\ref{eq:quotient-manifold-subspace-proj}). The canonical projection $\pi$ is, thus, obtained by the group action of Lie groups ${\rm GL}(r)$ and $\OG{r}$, the fiber spaces of the fixed-rank matrix factorizations. Hence, by the direct application of \cite[Theorem~9.16]{lee03a}, the quotient spaces of the matrix factorizations have the structure of smooth quotient manifolds and the map $\pi$ is a smooth submersion for each of the quotient spaces. Table \ref{tab:matrix_representations} shows the matrix representations of different fixed-rank matrix factorizations considered earlier in Section \ref{sec:quotient_spaces}.

\begin{table}
\begin{center} \scriptsize
\begin{tabular}{ p{2.0cm} | p{2.0cm} | p{4.0cm} | p{2.0cm} } 
& $\mat{W} = \mat{GH}^T$ & $\mat{W} = \mat{UBV}^T$ & $\mat{W} = \mat{UY}^T$ \\
\hline
& & & \\

Matrix $ $ $ $representation &  $(\mat{G}, \mat{H})$
 & $(\mat{U}, \mat{B}, \mat{V} )$ & $(\mat{U}, \mat{Y})$  \\ 
& & & \\

Total space $\overline{\mathcal W}$ &  $\mathbb{R}_*^{d_1 \times r} \times \mathbb{R}_*^{d_2 \times r}$
 & $\Stiefel{r}{d_1} \times \ConePD{r} \times \Stiefel{r}{d_2}$ & 
 $
 \begin{array}[t]{lll}
  \Stiefel{r}{d_1}\times\mathbb{R}_*^{d_2 \times r}\\
\end{array}  
 $ \\ 
 & & & \\

Group action &  $(\mat{GM}^{-1}, \mat{HM}^T)$
 & $(\mat{UO}, \mat{O}^T\mat{BO}, \mat{VO} )$ & $(\mat{UO}, \mat{YO})$  \\

& $\mat{M} \in  {\rm GL}(r)$& $\mat{O} \in \OG{r}$ &   $\mat{O} \in \OG{r}$ \\
 & & & \\
Quotient space ${\mathcal W}$ &  
$
\begin{array}[t]{lll}
\mathbb{R}_*^{d_1 \times r} \times \mathbb{R}_*^{d_2 \times r} \\
/  {\rm GL}(r) 
\end{array}
$

 & 
 $
 \begin{array}[t]{lll}
 \Stiefel{r}{d_1} \times \ConePD{r} \times \Stiefel{r}{d_2}\\
  / \OG{r} 
\end{array} 
 $ & 
 $
 \begin{array}[t]{lll} 
 \Stiefel{r}{d_1}\times \mathbb{R}_*^{d_2 \times r} \\
 / \OG{r}
\end{array} 
 $ \\ 

\hline
\end{tabular}
\end{center} 
\caption{Fixed-rank matrix factorizations and their quotient manifold representations. The action of Lie groups ${\rm GL}(r)$ and $\OG{r}$ make the quotient spaces, smooth quotient manifolds \cite[Theorem~9.16]{lee03a}.}
\label{tab:matrix_representations} 
\end{table}

\subsection{Tangent vector representation as horizontal lifts}\label{sec:horizontal_lifts}
Calculus on a manifold $\mathcal{W}$ is developed in the tangent space $T_{x} \mathcal{W}$, a vector space that can be considered as the linearization of the nonlinear space ${\mathcal W}$ at $x$. Since, the manifold $\mathcal{W}$ is an abstract space, the elements of  its tangent space $T_{x} \mathcal{W}$ at $x \in \mathcal{W}$ call for a  matrix representation in the total space $\overline{\mathcal{W}}$ at $\bar{x}$ that respects the equivalence relationship $\sim$. In other words, the matrix representation of $T_{x} \mathcal{W}$ should be restricted to the directions in the tangent space $T_{\bar x } \overline{\mathcal{W}}$ in the total space $\overline{\mathcal W}$ at ${\bar x}$ that do not induce a displacement along the equivalence class $[x]$. 

On the other hand, the tangent space at $\bar{x}$ of the total space $\overline{\mathcal W}$ admits a product structure, similar to the product structure of the total space. Because the total space is a product space of $\mathbb{R}_*^{d_1 \times r}$, $\Stiefel{r}{d_1}$ and $\ConePD{r}$, its tangent space $\overline{\mathcal{W}}$ at $\bar x$ embodies the product space of the tangent spaces of $\mathbb{R}_*^{d_1 \times r}$, $\Stiefel{r}{d_1}$ and $\ConePD{r}$, the characterizations of which are well-known. Refer \cite[Section~2.2]{edelman98a} or \cite[Example~3.5.2]{absil08a} for the characterization of the tangent space of $\Stiefel{r}{d_1}$. Similarly. the tangent spaces of $\mathbb{R}_*^{d_1 \times r}$ and $\ConePD{r}$ are $\mathbb{R}^{d_1 \times r}$ (Euclidean space) and $\symmetric{r}$ (the set of symmetric $r\times r$ matrices) respectively. 

The matrix representation of a tangent vector at $x \in \mathcal{W}$ relies on the decomposition of $T_{\bar x} \overline{\mathcal W}$ into complementary subspaces, \emph{vertical} and \emph{horizontal} subspaces. The vertical space $\mathcal{V}_{\bar x} \overline{\mathcal W}$ is the tangent space of the equivalence class $T_{\bar x} \pi^{-1}(x)$. The horizontal space $\mathcal{H}_{\bar x} \overline{\mathcal W}$, the complementary space of $\mathcal{V}_{\bar x} \overline{\mathcal W}$, then provides a valid matrix representation of the abstract tangent space $T_{x} \mathcal{W}$ \cite[Section~3.5.8]{absil08a}. The tangent vector $\bar{\xi}_{\bar{x}}\in\mathcal{H}_{\bar{x}}\overline{\mathcal{W}}$ is called the \emph{horizontal lift} of $\xi_{x}$ at $\bar{x}$. Refer to Figure \ref{fig:quotient_optimization} for a graphical illustration.

\begin{table}
\begin{center} \scriptsize
\begin{tabular}{ p{1.5cm} | p{3.5cm} | p{3.0cm} | p{2.0cm} } 
& $\mat{W} = \mat{GH}^T$ & $\mat{W} = \mat{UBV}^T$ & $\mat{W} = \mat{UY}^T$ \\
\hline
& & & \\

Tangent vectors in $\overline{\mathcal W}$&  
$
\begin{array}[t]{lll}
(\bar{\xi}_{\mat{G}}, \bar{\xi}_{\mat{H}} )  \in  \\
\mathbb{R}^{d_1 \times r} \times  \mathbb{R}^{d_2 \times r} 
\end{array}
$
 & 
 $
\begin{array}[t]{lll}
 & (\mat{Z}_{\mat U}, \mat{Z}_{\mat B}, \mat{Z}_{\mat V}) \in  \\
& \mathbb{R}^{d_1 \times r} \times \mathbb{R}^{r \times r}  \times  \mathbb{R}^{d_2 \times r}: \\

&  \mat{U}^T  \mat{Z}_\mat{U} + \mat{Z}_\mat{U} ^T \mat{U} = 0, \\

&   \mat{Z}_{\mat B}^T = \mat{Z}_{\mat B} , \\
& \mat{V}^T  \mat{Z}_\mat{V} + \mat{Z}_\mat{V} ^T \mat{V} = 0 \\

\end{array}
$ 
  & 
  $ 
\begin{array}[t]{lll}
 (\mat{Z}_{\mat U}, \mat{Z}_{\mat Y}) \in  \\
\mathbb{R}^{d_1 \times r}  \times  \mathbb{R}^{d_2 \times r}: \\

\mat{U}^T  \mat{Z}_\mat{U} + \mat{Z}_\mat{U} ^T \mat{U} = 0\\

\end{array}  
  $  \\ 
& & & \\

Metric 
$
\bar{g}_{\bar x} (\bar{\xi}_{\bar x},  \bar{\eta}_{\bar x}, )
$
&
 $
 \begin{array}[t]{lll}
  \trace ( (\mat{G}^T\mat{G})^{-1} \bar{\xi}^T_{\mat G} \bar{\eta}_{\mat G}  )  \\ 
 +  \trace ( (\mat{H}^T\mat{H})^{-1} \bar{\xi}^T_{\mat H} \bar{\eta}_{\mat H}  )
 \end{array}
 $
&  

$
 \begin{array}[t]{lll}
    \trace  (\bar{\xi}_{\mat U}^T  \bar{\eta}_{\mat U})   \\ 
+  \trace (  \mat{B}^{-1} \bar{\xi}_{\mat B} \mat{B}^{-1} \bar{\eta}_{\mat B}  ) \\
 + \trace  (\bar{\xi}_{\mat V}^T  \bar{\eta}_{\mat V})   \\ 
 \end{array}
 $

&

$
 \begin{array}[t]{lll}
    \trace  (\bar{\xi}_{\mat U}^T  \bar{\eta}_{\mat U})   \\ 
 + \trace  ( (\mat{Y}^T\mat{Y})^{-1}  \bar{\xi}_{\mat Y}^T  \bar{\eta}_{\mat Y})   
 \end{array}
 $

 \\
& & & \\

Vertical tangent vectors 
& 
$
\begin{array}[t]{lll}
  (   -\mat{G}\mat{\Lambda} , \mat{H}\mat{\Lambda} ^T   )  : \\
  \mat{\Lambda} \in \mathbb{R}^{r\times r} 
\end{array}
$
& 
$
\begin{array}[t]{lll}
   (   \mat{U}\mat{\Omega} ,   \mat{B}\mat{\Omega} -  \mat{\Omega}\mat{B},   \mat{V}\mat{\Omega}   ):  \\ \mat{\Omega}^T = - \mat{\Omega} 
\end{array}
$
& 
$
\begin{array}[t]{lll}
   (   \mat{U}\mat{\Omega} ,   \mat{Y}\mat{\Omega}   ):  \\ \mat{\Omega}^T = - \mat{\Omega} 
\end{array}
$
\\

& & & \\

Horizontal tangent vectors
& 
$
\begin{array}[t]{lll}
  \left ( \bar{ \zeta}_{\mat{G}} , \bar{\zeta}_{\mat{H}}   \right )
  \in \mathbb{R}^{d_1 \times r}  \times \mathbb{R}^{d_2 \times r} :\\
    \bar{\zeta}_{\mat{G}}^T \mat{G} \mat{H}^T \mat{H} 
   = \mat{G}^T \mat{G} \mat{H}^T \bar{\zeta}_\mat{H}
 \end{array}
$
& 
$
\begin{array}[t]{lll}
   (  \zeta_{\mat{U}} , \zeta_\mat{B}, \zeta_{\mat{V}}   )  \in  T_{\bar{x}} \overline{\mathcal{W}}: \\
  (\zeta_{\mat{U}}^T\mat{U} + \mat{B}^{-1}\zeta_{\mat B}  - \zeta_{\mat B} \mat{B}^{-1} \\ 
  +   \zeta_{\mat{V}}^T\mat{V} ) \rm{\  is\ symmetric } 
 
 \end{array}
$
& 
$
\begin{array}[t]{lll}
   (  \zeta_{\mat{U}} , \zeta_\mat{Y} )  \in  T_{\bar{x}} \overline{\mathcal{W}} : \\
\zeta_{\mat{U}}^T \mat{U} + (\mat{Y}^T \mat{Y} )^{-1}\zeta_{\mat{Y}}^T\mat{Y}\ \\ {\rm is\ symmetric} 
 \end{array}
$
\\

& & & \\

\hline
\end{tabular}
\end{center} 
\caption{Matrix representations of tangent vectors. The tangent space $T_{\bar x} \overline{\mathcal W}$ in the total space is decomposed into  orthogonal subspaces, the vertical space $\mathcal{V}_{\bar x} \overline{\mathcal{W}}$ and the horizontal space $\mathcal{H}_{\bar x} \overline{\mathcal{W}}$. The Riemannian metric is chosen by picking the natural metric for each of the space, $\mathbb{R}_*^{d_1 \times r}$ \cite[Example~3.6.4]{absil08a}, $\Stiefel{r}{d_1}$ \cite[Example~3.6.2]{absil08a} and $\ConePD{r}$ \cite[Section~6.1]{bhatia07a}. The Riemannian metric $\bar{g}_{\bar x}$ makes the matrix representation of the abstract tangent space $T_{ x} {\mathcal W}$ unique in terms of the horizontal space $\mathcal{H}_{\bar x} \overline{\mathcal{W}}$.}
\label{tab:spaces} 
\end{table}

\begin{table}
\begin{center} \scriptsize
\begin{tabular}{ p{2.2cm} | p{3.2cm} | p{4.0cm} | p{3.0cm} } 
& $\mat{W} = \mat{GH}^T$ & $\mat{W} = \mat{UBV}^T$ & $\mat{W} = \mat{UY}^T$ \\
\hline
& & & \\

Matrix representation of the ambient space 
&
$
\begin{array}[t]{lll}
(\mat{Z}_{\mat{G}}, \mat{Z}_{\mat{H}}) \in \\
\mathbb{R}^{d_1 \times r} \times \mathbb{R}^{d_2 \times r}
\end{array}
$
&
$
\begin{array}[t]{lll}
(\mat{Z}_{\mat{U}}, \mat{Z}_{\mat{B}}, \mat{Z}_{\mat{V}}) \in \\
\mathbb{R}^{d_1 \times r} \times \mathbb{R}^{r \times r } \times \mathbb{R}^{d_2 \times r}
\end{array}
$
&
$
\begin{array}[t]{lll}
(\mat{Z}_{\mat{U}}, \mat{Z}_{\mat{Y}}) \in \\
\mathbb{R}^{d_1 \times r} \times \mathbb{R}^{d_2 \times r}
\end{array}
$
\\
\hline

\multicolumn{4}{ c }{ {\huge $\downarrow$} { $ \Psi_{\bar x}$} } \\

\hline
Projection onto $T_{\bar x} \overline{\mathcal W}$
&  
$(\mat{Z}_{\mat{G}}, \mat{Z}_{\mat{H}})$
 & 
 $
\begin{array}[t]{ll}
 ( \mat{Z}_\mat{U} - \mat{U} \Sym(\mat{U}^T \mat{Z}_\mat{U}), \\
 \Sym(\mat{Z}_{\mat{B}} ), \\
\mat{Z}_\mat{V} - \mat{V} \Sym(\mat{V}^T \mat{Z}_\mat{V}))
  \end{array}
$
 &
$
\begin{array}[t]{ll}
 ( \mat{Z}_\mat{U} - \mat{U} \Sym(\mat{U}^T \mat{Z}_\mat{U}), \\
\mat{Z}_\mat{Y} )
  \end{array}
$ 
 
 \\

& & & \\

\hline
\multicolumn{4}{ c }{ {\huge $\downarrow$} { $ \Pi_{\bar x}$} } \\

\hline

& & & \\

Projection of a tangent vector $\bar{\eta}_{\bar x} \in T_{\bar x} \overline{\mathcal W}$ onto $\mathcal{H}_{\bar x} \overline{\mathcal W}$
&  
$
\begin{array}[t]{ll}
 ( \bar{\eta}_{\mat{U}} + \mat{G}\mat{\Lambda} , \bar{\eta}_{\mat{H}} - \mat{H\Lambda}^T )
 
 \\
 \\
 \\
\end{array}
$  
where $\mat{\Lambda}$ is the unique solution to the Lyapunov equation
$
\begin{array}[t]{lll}
\mat{\Lambda}^T (\mat{G}^T \mat{G}) (\mat{H}^ T \mat{H} ) \\ 
+   ( \mat{G}^T \mat{G}) (\mat{H}^ T \mat{H}       ) \mat{\Lambda}^T  = \\
  (\mat{G}^T \mat{G}) \mat{H}^ T\bar{ \eta}_\mat{H}  \\
  - \bar{\eta}_{\mat{G}}^T  \mat{G} (\mat{H}^ T \mat{H})
\end{array}
$
 &

$
\begin{array}[t]{ll}
 ( \bar{\eta}_{\mat{U}} - \mat{U}\mat{\Omega} , \bar{\eta}_{\mat{B}} - (\mat{B\Omega} - \mat{\Omega B}),\\
  \bar{\eta}_{\mat{V}} - \mat{V}\mat{\Omega}   )
  \\
  \\
\end{array}  
$  
where $\mat{\Omega}$ is the  unique solution to the Lyapunov equation
$
\begin{array}[t]{llll}
\\
\mat{\Omega} \mat{B}^2 + \mat{B}^2 \mat{\Omega} = \\
\mat{B} ( \Skew(\mat{U}^T \bar{\eta}_{\mat{U}})\\
 - 2 \Skew(\mat{B}^{-1} \bar{\eta}_{\mat {B}}) \\
+ \Skew(\mat{V}^T \bar{\eta}_{\mat{V}})    ) \mat{B}
\end{array}
$

 &
$
\begin{array}[t]{ll}
( \bar{\eta}_{\mat{U}} - \mat{U}\mat{\Omega} ,\\ \bar{\eta}_{\mat{Y}} - \mat{Y\Omega} )
\\
  \\
\end{array}
$  where $\mat{\Omega}$ is the unique solution to 
$
\begin{array}[t]{llll}
( \mat{Y}^T\mat{Y} ) \widetilde{{\mat \Omega}}  +  \widetilde{ {\mat \Omega}}  ( \mat{Y}^T\mat{Y} )  \\
 =   2 \Skew(( \mat{Y}^T\mat{Y} )(\mat{U}^T\bar{\eta}_{\mat{U}})( \mat{Y}^T\mat{Y} ) )   \\
- 2\Skew((\bar{\eta}_{\mat{Y}}^T \mat{Y})( \mat{Y}^T\mat{Y} ))\\
\\
{\rm and} \\
\\
(\mat{Y}^T\mat{Y})\mat{\Omega} + \mat{\Omega}(\mat{Y}^T\mat{Y}) \\
= \widetilde{{\mat \Omega}} 
\end{array}
$

 \\

& & & \\
\hline
\end{tabular}
\end{center} 
\caption{The matrix representations of the projection operators $\Psi_{\bar x}$ and $\Pi_{\bar x}$. $\Psi_{\bar x}$ projects a matrix in the Euclidean space onto the tangent space $T_{\bar x}\overline{\mathcal W}$. $\Pi_{\bar x}$ extracts the horizontal component of a tangent vector $\bar{\xi}_{\bar x}$. Here the operators $\Sym(\cdot)$ and $\Skew(\cdot)$ extract the symmetric and skew-symmetric parts of a square matrix and are defined as $\Sym(\mat{A}) = \frac{\mat{A} + \mat{A}^T}{2}$ and  $\Skew(\mat{A}) = \frac{\mat{A}^T - \mat{A}}{2}$ for any square matrix $\mat{A}$.}
\label{tab:projections} 
\end{table}

A metric $\bar{g}_{\bar{x}}(\bar{\xi}_{\bar{x}},\bar{\zeta}_{\bar{x}})$ on the total space defines a valid Riemannian metric $g_x$ on the quotient manifold if
\begin{equation}\label{eq:metric}
	g_{x}(\xi_{x},\zeta_{x}):=\bar{g}_{\bar{x}}(\bar{\xi}_{\bar{x}},\bar{\zeta}_{\bar{x}})
\end{equation}
where $\xi_x$ and $\zeta_x$ are the tangent vectors in $T_x \mathcal{W}$ and $\xi_{\bar{x}}$ and $\zeta_{\bar{x}}$ are their horizontal lifts in $\mathcal{H}_{\bar{x}} \overline{\mathcal{W}}$. The product structure of the total space $\overline{\mathcal{W}}$ again allows us to define a valid Riemannian metric by picking the natural metric for $\mathbb{R}_*^{d_1 \times r}$ \cite[Example~3.6.4]{absil08a}, $\Stiefel{r}{d_1}$ \cite[Example~3.6.2]{absil08a} and $\ConePD{r}$ \cite[Section~6.1]{bhatia07a}. Endowed with this Riemannian metric, $\mathcal{W}$ is called a \emph{Riemannian quotient manifold} of $\overline{\mathcal{W}}$ and the quotient map $\pi: \overline{\mathcal W} \rightarrow \mathcal{W}$ is a \emph{Riemannian submersion} \cite[Section~3.6.2]{absil08a}. Once $T_{\bar{x}}\overline{\mathcal{W}}$ is endowed with a horizontal distribution $\mathcal{H}_{\bar{x}}\overline{\mathcal{W}}$ (as a result of the Riemannian metric), a given tangent vector $\xi_{x} \in T_{x}\mathcal{W}$ at $x$ on the quotient manifold $\mathcal{W}$ is uniquely represented by the tangent vector $\bar{\xi}_{\bar{x}}\in\mathcal{H}_{\bar{x}}\overline{\mathcal{W}}$ in the total space $\overline{\mathcal{W}}$ that satisfies ${\rm D} \pi(\bar{x})[\bar{\xi}_{\bar x} ] = \xi_{x}$. The matrix characterizations of the $T_{\bar x} \overline{\mathcal W}$, $\mathcal{V}_{\bar x} \overline{\mathcal W}$ and $\mathcal{H}_{\bar x} \overline{\mathcal W}$ and the Riemannian metric $\bar{g}_{\bar x}$  for the three considered matrix factorizations are given in Table \ref{tab:spaces}.

\change{Table \ref{tab:projections} summarizes the concrete matrix operations involved in computing horizontal vectors. Starting from an arbitrary matrix (with appropriate dimensions), two linear projections are needed: the first projection $\Psi_{\bar x}$ is onto the tangent space of the total space, while the second projection $\Pi_{\bar x}$ is onto the horizontal subspace. Note that all matrix operations are linear in the original matrix dimensions ($d_1$ or $d_2$). This is critical for the computational efficiency of the matrix algorithms. }

\subsection{Retractions from the tangent space to the manifold}\label{sec:retraction}
An iterative optimization algorithm involves computing a (e.g. gradient) search direction and then ``moving in that direction''. The default option on a Riemannian manifold is to move along geodesics, leading to the definition of the exponential map (see e.g  \cite[Chapter~20]{lee03a}). Because the calculation of the exponential map can be computationally  demanding, it is customary in the context of manifold optimization to relax the constraint of moving along geodesics. The exponential map is then relaxed to a {\it retraction}, which is any map  $R_{\bar x}: \mathcal{H}_{\bar x} \overline{\mathcal{W}} \rightarrow \overline{\mathcal{W}} $ that locally approximates the exponential map  on the manifold \cite[Definition~4.1.1]{absil08a}.
A natural update on the manifold is, thus, based on the update formula
\begin{equation}\label{eq:retraction}
	\bar{x}_{+} = R_{\bar{x}}( \bar{\xi}_{\bar{x}})
\end{equation}
where $\bar{\xi}_{\bar x} \in \mathcal{H}_{\bar x} \overline{\mathcal W}$ is a search direction and $\bar{x}_+ \in \overline{\mathcal W}$. See Figure \ref{fig:quotient_optimization} for a graphical view. Due to the product structure of the total space, a retraction is obtained by combining the retraction updates on $\mathbb{R}_*^{d_1 \times r}$ \cite[Example~4.1.5]{absil08a}, $\Stiefel{r}{d_1}$ \cite[Example~4.1.3]{absil08a} and $\ConePD{r}$ \cite[Theorem~6.1.6]{bhatia07a}. Note that the retraction on the positive definite cone is the exponential mapping with the natural metric \cite[Theorem~6.1.6]{bhatia07a}. The cartesian product of the retractions also defines a valid retraction on the quotient manifold $\mathcal{W}$ \cite[Proposition~4.1.3]{absil08a}. The retractions for the fixed-rank matrix factorizations are presented in Table \ref{tab:retraction}. The reader will notice that the matrix computations involved are again linear in the matrix dimensions $d_1$ and $d_2$.

\begin{table}
\begin{center} \scriptsize
\begin{tabular}{ p{2.0cm} | p{2.0cm} | p{3.3cm} | p{2.0cm} } 
& $\mat{W} = \mat{GH}^T$ & $\mat{W} = \mat{UBV}^T$ & $\mat{W} = \mat{UY}^T$ \\
\hline
& & & \\
Retraction $R_{\bar x}({\bar \xi}_{\bar x})$ that maps a

horizontal 

vector $\bar{\xi}_{\bar x}$ onto $\overline{\mathcal W}$

&
$
\begin{array}[t]{ll}
(\mat{G}  + \bar{\xi}_{\mat G} ,
\\
 \mat{H} + \bar{\xi}_{\mat H}) \\
\end{array}
$

&
$
\begin{array}[t]{ll}
(\rm{uf}(\mat{U}  + \bar{\xi}_{\mat U}), \\
 \mat{B}^{\frac{1}{2}} \rm{exp}   (    \mat{B}^{- \frac{1}{2}}  \xi_{\mat B}  \mat{B}^{- \frac{1}{2}}   )   \mat{B}^{\frac{1}{2}}, \\

\rm{uf}(\mat{V}  + \bar{\xi}_{\mat V}) ) \\

\end{array}
$
&
$
\begin{array}[t]{ll}
(\rm{uf}(\mat{U}  + \bar{\xi}_{\mat U}) ,
\\
 \mat{Y} + \bar{\xi}_{\mat Y} )\\
\end{array}
$
\\

\hline
\end{tabular}
\end{center} 
\caption{Retraction $R_{\bar x}(\cdot)$ maps a horizontal vector $\bar{\xi}_{\bar x}$ on the manifold $\overline{\mathcal W}$. It provides a computationally efficient way to move on the manifold while approximating the geodesics. ${\rm uf}(\cdot)$ extracts the orthogonal factor of a full column rank matrix $\mat{D}$, i.e., ${\rm uf}(\mat{D}) = \mat{D} (\mat{D}^T\mat{D})^{-1/2}$ and ${\rm exp}(\cdot)$ is the matrix exponential operator. }
\label{tab:retraction} 
\end{table}

\subsection{Gradient and Hessian in Riemannian submersions} 
\label{sec:gradient_Hessian}
The choice of the metric (\ref{eq:metric}), which is invariant along the equivalence class $[\bar{x}]$, and of the horizontal space (as the orthogonal complement of $\mathcal{V}_{\bar x} \overline{\mathcal W}$ in the sense of the Riemannian metric) turns the quotient manifold $\mathcal{W}$ into a Riemannian submersion of $(\overline{\mathcal{W}}, \bar{g})$ \cite[Section~3.6.2]{absil08a}. As shown in \cite{absil08a}, this special construction allows for a convenient matrix representation of the gradient \cite[Section~3.6.2]{absil08a} and the Hessian \cite[Proposition~5.3.3]{absil08a} on the abstract manifold $\mathcal{W}$. 

Any smooth cost function $\bar{\phi} : \overline{\mathcal W} \rightarrow \mathbb{R}$ which is invariant along the fibers induces a corresponding smooth function $\phi$ on the quotient manifold $\mathcal{W}$. The Riemannian gradient of $\phi$ is uniquely represented by its horizontal lift in $\overline{\mathcal{W}}$ which has the matrix representation
\begin{equation}\label{eq:Riemannian_gradient}
\overline { {\grad}_x \phi} = \grad_{\bar{x}} \bar{\phi}.
\end{equation}
It should be emphasized that $\grad_{\bar{x}} \bar{\phi}$ is in the the tangent space $T_{\bar{x}} \overline{\mathcal{W}}$. However, due to invariance of the cost along the equivalence class $[\bar{x}]$, $\grad_{\bar{x}} \bar{\phi}$ also belongs to the horizontal space $\mathcal{H}_{\bar{x}} \overline{\mathcal{W}}$ and hence, the equality in (\ref{eq:Riemannian_gradient}) \cite[Section~3.6.2]{absil08a}. The matrix expression of $\grad_{\bar{x}} \bar{\phi}$ in the total space $\overline{\mathcal{W}}$ at a point $\bar{x}$ is obtained from its definition: it is the unique element of $T_{\bar x} \overline{\mathcal W}$ that satisfies ${\rm D} \bar{\phi}[\eta_{\bar x}] = \bar{g}_{\bar{x}}(     \grad_{\bar x} \bar{\phi},   \eta_{\bar{x}}  )$ for all $ \eta_{\bar x} \in T_{\bar x} \overline{\mathcal W}$ \cite[Equation~3.31]{absil08a}. ${\rm D} \bar{\phi}[\eta_{\bar x}]$ is the standard Euclidean directional derivative of $\bar{\phi}$ in the direction $\eta_{\bar x}$ and $\bar{g}_{\bar x}$ is the Riemannian metric. This definition leads to the matrix representations of the Riemannian gradient in Table \ref{tab:gradient_Hessian}.

\begin{table}
\begin{center} \scriptsize
\begin{tabular}{ p{1.3cm} | p{3.6cm} | p{3.5cm} | p{3.5cm} } 
& $\mat{W} = \mat{GH}^T$ & $\mat{W} = \mat{UBV}^T$ & $\mat{W} = \mat{UY}^T$ \\
\hline
& & & \\

Riemannian gradient $\grad_{\bar x} \bar{\phi}$
&
First compute the partial derivatives

$
\begin{array}[t]{lll}
& (\bar{\phi}_{\mat{G}}, \bar{\phi}_{\mat{H}}) \in \\
& \mathbb{R}^{d_1 \times r} \times \mathbb{R}^{d_2 \times r}
\\
\\
\end{array}
$

and then perform the operation

$
\begin{array}[t]{lll}
\\
(\bar{\phi}_{\mat{G}}  {\mat G}^T{\mat G}, \bar{\phi}_{\mat{H}}  {\mat H}^T{\mat H})  
\\

\end{array}
$

&
First compute the partial derivatives

$
\begin{array}[t]{lll}

(\bar{\phi}_{\mat{U}}, \bar{\phi}_{\mat{B}}, \bar{\phi}_{\mat{V}}) \in \\
\mathbb{R}^{d_1 \times r} \times \mathbb{R}^{r \times r } \times \mathbb{R}^{d_2 \times r}
\\
\\
\end{array}
$

and then perform the operation
 
$
\begin{array}[t]{lll}
\\
(\bar{\phi}_{\mat{U}} - \mat{U}^T\Sym(\mat{U}^T \bar{\phi}_{\mat{U}}), \\

\mat{B} \Sym(\bar{\phi}_{\mat{B}}) \mat{B}, \\

\bar{\phi}_{\mat{V}} - \mat{V}^T\Sym(\mat{V}^T \bar{\phi}_{\mat{V}}) )
\end{array}
$ 
 
&

First compute the partial derivatives

$
\begin{array}[t]{lll}
(\bar{\phi}_{\mat{U}}, \bar{\phi}_{\mat{Y}}) \in \\
\mathbb{R}^{d_1 \times r} \times \mathbb{R}^{d_2 \times r}
\\
\\
\end{array}
$

and then perform the operation

$
\begin{array}[t]{lll}
\\
(\bar{\phi}_{\mat{U}} - \mat{U}^T\Sym(\mat{U}^T \bar{\phi}_{\mat{U}}), \\

\bar{\phi}_{\mat{Y}}  {\mat Y}^T{\mat Y}) 
\end{array}
$

\\

& & & \\

& & & \\

Riemannian 

connection     

$\overline{\rc}_{\bar{\xi}_{\bar x}} \bar{\eta}_{\bar x}$

&

$
\begin{array}[t]{lll}
\Psi_{\bar x} (\D \bar{\eta}_{\bar x}  [\bar{\xi}_{\bar x}] +  (\mat{A}_{\mat G}, \mat{A}_{\mat H}) )\\
 \\
 \\
{\rm where} \\

\mat{A}_{\mat G} = \\
  - \bar{\eta}_\mat{G} { ( \mat{G}^T \mat{G}  )}^{-1}  \Sym{ ( \mat{G}^T \bar{\xi}_\mat{G}  )}   \\
  -\bar{\xi}_\mat{G} {  ( \mat{G}^T \mat{G}  )}^{-1} \Sym{ ( \mat{G}^T \bar{\eta}_\mat{G}  )}   \\
   + \mat{G} {  ( \mat{G}^T \mat{G}  )}^{-1} \Sym{ ( \bar{\eta}_\mat{G}^T \bar{\xi}_\mat{G} )}
 ), \\

\\
\mat{A}_{\mat H} =  \\
 - \bar{\eta}_\mat{H} { ( \mat{H}^T \mat{H}  )}^{-1}  \Sym{ ( \mat{H}^T \bar{\xi}_\mat{H}  )}   \\
  -\bar{\xi}_\mat{H} {  ( \mat{H}^T \mat{H}  )}^{-1} \Sym{ ( \mat{H}^T \bar{\eta}_\mat{H}  )}   \\
   + \mat{H} {  ( \mat{H}^T \mat{H}  )}^{-1} \Sym{ ( \bar{\eta}_\mat{H}^T \bar{\xi}_\mat{H} )}
 )
\end{array}
$

&

$
\begin{array}[t]{lll}
 \Psi_{\bar{x}}(\D \bar{\eta}_{\bar x}  [\bar{\xi}_{\bar x}]  \\
 + (\mat{A}_{\mat U}, \mat{A}_{\mat B}, \mat{A}_{\mat V}) ) \\

\\
{\rm where} \\

 \mat{A}_{\mat U} =   -   \bar{\xi}_{\mat{U}}\Sym(\mat{U}^T \bar{\eta}_{\mat{U}}),\\
 
\mat{A}_{\mat B} =  - \Sym(\xi_{\mat{B}} \mat{B}^{-1} \eta_{\mat{B}}), \\
 \mat{A}_{\mat V} = -  \bar{\xi}_{\mat{V}}\Sym(\mat{V}^T \bar{\eta}_{\mat{V}})
 
\end{array}
$

&
$
\begin{array}[t]{lll}
   \Psi_{\bar{x}}(\D \bar{\eta}_{\bar x}  [\bar{\xi}_{\bar x}]  \\
   +  (\mat{A}_{\mat U}, \mat{A}_{\mat Y}) ) \\

\\
{\rm where} \\

\mat{A}_{\mat U} =  - \bar{\xi}_{\mat{U}}\Sym(\mat{U}^T \bar{\eta}_{\mat{U}}),
\\

\\
\mat{A}_{\mat Y} = \\
 - \bar{\eta}_\mat{Y} { ( \mat{Y}^T \mat{Y}  )}^{-1}  \Sym{ ( \mat{Y}^T \bar{\xi}_\mat{Y}  )}   \\
  -\bar{\xi}_\mat{Y} {  ( \mat{Y}^T \mat{Y}  )}^{-1} \Sym{ ( \mat{Y}^T \bar{\eta}_\mat{Y}  )}   \\
   +\  \mat{Y} {  ( \mat{Y}^T \mat{Y}  )}^{-1} \Sym{ ( \bar{\eta}_\mat{Y}^T \bar{\xi}_\mat{Y} )}
 
\end{array}
$

\\
& & & \\

\hline

\end{tabular}
\end{center} 
\caption{The Riemannian gradient of the function $\bar{\phi}$ and the Riemannian connection at $\bar{x}$ in total space $\overline{\mathcal W}$. The matrix representations of their counterparts on the Riemannian quotient manifold ${\mathcal W}$ are given by (\ref{eq:Riemannian_gradient}) and (\ref{eq:Riemannian_connection}). Here $\D \bar{\eta}_{\bar x}  [\bar{\xi}_{\bar x}] $ is the standard Euclidean directional derivative of the vector field ${\bar \eta}_{\bar x}$ in the direction ${\bar \xi}_{\bar x}$, i.e., $\D \bar{\eta}_{\bar x}  [\bar{\xi}_{\bar x}]  = \lim_{t \rightarrow 0^+} \frac{\bar{\eta}_{\bar{x} + t \bar{\xi}_{\bar x} } - \bar{\eta}_{\bar x}}{t}$. The projection operator $\Psi_{\bar x}$ maps an arbitrary matrix in the Euclidean space on the tangent space $T_{\bar x} \overline{\mathcal W}$ and is defined in Table \ref{tab:projections}. 
}
\label{tab:gradient_Hessian} 
\end{table}

In addition to the gradient, \change{any optimization algorithm that makes use of second-order information also requires the directional derivative of the gradient along a search direction. This involves the choice of an \emph{affine connection} $\rc$ on the manifold. The affine connection provides a definition for  the \emph{covariant derivative} of vector field $\eta_x$ with respect to the vector field $\xi_x$, denoted by  $\rc _{\xi_x} \eta_x $. Imposing an additional compatibility condition with the metric fixes the so-called  \emph{Riemannian} connection} which is always unique \cite[Theorem~5.3.1 and Section~5.2]{absil08a}. The Riemannian connection $\rc_{\xi_x} \eta_x$ on the quotient manifold $\mathcal{W}$ is uniquely represented in terms of the Riemannian connection in the total space $\overline{\mathcal{W}}$, $\overline{\rc}_{\bar{\xi}_{\bar x}} \bar{\eta}_{\bar x}$ \cite[Proposition~5.3.3]{absil08a} which is 
\begin{equation} \label{eq:Riemannian_connection}
\overline { {\rc}_{\xi _x} {\eta _x}} = \Pi_{\bar{x}} (\overline{\rc}_{\bar{\xi}_{\bar x}} \bar{\eta}_{\bar x})
\end{equation}
where $\xi_{x}$ and $\eta_x$ are vector fields in $\mathcal{W}$ and $\bar{\xi}_{\bar x}$ and $\bar{\eta}_{\bar x}$ are their horizontal lifts in $\overline{\mathcal{W}}$. Here $\Pi_{\bar x}$ is the projection operator that projects a tangent vector in $T_{\bar x} \overline{\mathcal W}$ onto the horizontal space $\mathcal{H}_{\bar x} \overline{\mathcal W}$ as defined in Table \ref{tab:projections}. In this case as well, the Riemannian connection $\overline{\rc}_{\bar{\xi}_{\bar x}} \bar{\eta}_{\bar x}$ on the total space $\overline{\mathcal{W}}$ has well-known expression owing to the product structure.

The Riemannian connection on the Stiefel manifold $\Stiefel{r}{d_1}$ is derived in \cite[Example $4.3.6$]{journee09a}. The Riemannian conenction on $\mathbb{R}_*^{d_1 \times r}$ and on the set of positive definite matrices $\ConePD{r}$ with their natural metrics are derived in \cite[Appendix B]{meyer11b}. Finally, the Riemannian connection on the total space is given by the cartesian product of the individual connections. In Table \ref{tab:gradient_Hessian} we give the final matrix expressions. The directional derivative of the Riemannian gradient in the direction $\xi_x$ is called the Riemannian Hessian $\hess_{x} \phi(x) [\xi _x]$ which is now directly given in terms of the Riemannian connection $\rc$. The horizontal lift of the Riemannian Hessian in ${\mathcal{W}}$ has, thus, the following matrix expression
\begin{equation}\label{eq:Riemannian_hessian}
\overline{\hess_{x} \phi(x) [\xi _x]}= \Pi_{\bar{x}}(  \overline{\rc}_{\bar{\xi}_{\bar x}} \overline{ \grad_{ x} \phi}   ).
\end{equation}
for any $\xi_x \in T_x \mathcal{W}$ and its horizontal lift $\bar{\xi}_{\bar x} \in \mathcal{H}_{\bar{x}} \overline{\mathcal{W}}$.

\section{Two optimization algorithms}\label{sec:algorithms}

\change{For the sake of illustration, we consider two basic optimization schemes in this paper: the (steepest) gradient descent algorithm, as a representative of first-order algorithms, and the Riemannian trust-region scheme, as a representative of second-order algorithms. Both schemes} can be easily implemented using the notions developed in the previous section. In particular, Table \ref{tab:projections} to \ref{tab:gradient_Hessian} give all the necessary ingredients for optimizing a smooth cost function ${\phi}: {\mathcal W} \rightarrow \mathbb{R}$ on the Riemannian quotient manifold of fixed-rank matrix factorizations.

\subsection{Gradient descent algorithm}\label{sec:gradient_descent}
For the gradient descent scheme we implement \cite[Algorithm~1]{absil08a} where at each iteration we move along the negative Riemannian gradient (see Table \ref{tab:gradient_Hessian}) direction by taking a step (\ref{eq:retraction}), and use the Armijo backtracking method \cite[Procedure~3.1]{nocedal06a} to compute an Armijo-optimal step-size satisfying the sufficient decrease condition \cite[Chapter~3]{nocedal06a}. The Riemannian gradient is the gradient of the cost function in the sense of the Riemannian metric proposed in Table \ref{tab:spaces}.

For computing an initial step-size, we use the information of the previous iteration by using the \emph{adaptive step-size update} procedure proposed below. The adaptive step-size update procedure is different from the initial step-size procedure described in \cite[Page~58]{nocedal06a}. This procedure is independent of the cost function evaluation and can be considered as a zero-order \emph{prediction heuristic}.

Let us assume that after the $t^{\rm th}$ iteration we know the initial step-size guess that was used $\hat{s}_t$, the Armijo-optimal step-size ${s}_t$ and the number of backtracking line-searches $j_t$ required to obtain the Armijo-optimal step-size ${s}_t$. The procedure is then,
\begin{equation}\label{eq:adaptive_stepsize}
\begin{array}[t]{lll}
{\rm Given:\ }&\hat{s}_t\  {\rm (initial \ step-size \ guess\ for\ iteration}\ t\ {\rm)} \\
                     & j_t \ {\rm (number\ of\ backtracking\ line-searches\ required\ at\ iteration}\ t{\rm )} \ {\rm and }  \\
                    &s_t\  {\rm (Armijo-optimal \ step-size) \ at \ iteration\  }  t. \\     
                    \\             
{\rm Then:\ } &  {\rm the\ initial\ step-size\ guess\ at\  iteration\ } t+1 \\
& {\rm is\ given\ by\ the\ update}  \\
& \hat{s}_{t+1} =  \left \{  
\begin{array}{ll}
2\hat{s}_t, &j_t = 0 \\
2{s}_t, & j_t = 1 \\
2s_t, & j_t \geq 2. 
\end{array}
 \right.               
\end{array}
\end{equation}
Here $s_0 (= {\hat s}_0)$ is the initial step-size guess provided by the user and $j_0 = 0$. \change{This procedure  keeps the number of line-searches close to $1$ on average, that is, $\mathbb{E}_t(j_t) \approx 1$, assuming that the optimal step-size does not vary too much with iterations.  An alternative is to choose any convex combination  of the following updates:}
\begin{equation*}
\hat{s}_{t+1}  = 
\left \{
\begin{array}{llll}
\underline{ {\rm update\ }1} & & \underline{{\rm update\ }2}  &\\
2\hat{s}_t  &   & 2\hat{s}_t , & j_t = 0 \\
2{s}_t  & &  1{s}_t, &  j_t = 1 \\
1s_t  &   & 2s_t,  &  j_t \geq 2.

\end{array}
\right.         
\end{equation*}

\subsection{Riemannian trust-region algorithm}\label{sec:trust_region}

The second optimization scheme we consider, is the Riemannian trust-region scheme. Analogous to trust-region algorithms in the Euclidean space \cite[Chapter~4]{nocedal06a}, trust-region algorithms on a Riemannian quotient manifold with guaranteed quadratic rate convergence have been proposed in \cite[Chapter~7]{absil08a}. Similar to the Euclidean case, at each iteration we solve the \emph{trust-region sub-problem} on the quotient manifold $\mathcal{W}$. The trust-region sub-problem is formulated as the minimization of the locally-quadratic model of the cost function, say $\phi : \mathcal{W} \rightarrow \mathbb{R}$ at $x\in \mathcal{W}$
\begin{equation}\label{eq:TR_subproblem}
\begin{array}{ll}
\min\limits_{\xi _x \in T_{x} \mathcal{W}} \quad & \phi (x) +  g_{x} (  \xi _x,   \grad _x \phi(x) ) + \frac{1}{2}  g_{x} (  \xi _x, \hess_x \phi(x) [\xi _x]    ) \\
\subject & {g}_{x}  ( \xi _x , \xi _x ) \leq \Delta ^2,
\end{array}
\end{equation}
where $\Delta$ is the trust-region radius, $g_x$ is the Riemannian metric; and $\grad _x \phi$ and $\hess_x \phi$ are the Riemannian gradient and Riemannian Hessian on the quotient manifold $\mathcal{W}$ (see Section \ref{sec:gradient_Hessian} and Table \ref{tab:gradient_Hessian}). The Riemannian gradient is the gradient of the cost function in the sense of the Riemannian metric $g_x$ and the Riemannian Hessian is given by the Riemannian connection. Computationally, the problem is horizontally lifted to the horizontal space $\mathcal{H}_{\bar{x}} \mathcal{W}$ \cite[Section~7.2.2]{absil08a} where we have the matrix representations of the Riemannian gradient and Riemannian Hessian (Table \ref{tab:gradient_Hessian}). Solving the above trust-region sub-problem leads to a direction $\bar{\xi}$ that minimizes the quadratic model. Depending on whether the decrease of the cost function is sufficient or not, the potential iterate is accepted or rejected. 

In particular, we implement the Riemannian trust-region algorithm \cite[Algorithm~10]{absil08a} using the generic solver GenRTR \cite{genrtr}. The trust-region sub-problem is solved using the \emph{truncated conjugate gradient} method \cite[Algorithm~11]{absil08a} which is does not require inverting the Hessian. The stopping criterion for the sub-problem is based on \cite[~(7.10)]{absil08a}, i.e.,
\[ 
\| r_{t+1} \| \leq \|r_0 \| \min (\|r_0 \|^{\theta}, \kappa)
\]
where $r_t$ is the residual of the sub-problem at $t^{\rm th}$ iteration of the truncated conjugate gradient method. The parameters $\theta$ and $\kappa$ are set to $1$ and $0.1$ as suggested in \cite[Section~7.5]{absil08a}. The parameter $\theta = 1$ ensures that we seek a quadratic rate of convergence near the minimum. 

\subsection{Numerical complexity}\label{sec:numerical_complexity}
The numerical complexity of manifold-based optimization methods depends on the computational cost of the components listed in Table \ref{tab:projections} to \ref{tab:gradient_Hessian} and the Riemannian metric ${\bar g}_{\bar x}$ presented in Table \ref{tab:spaces}. The computational cost of these ingredients are shown below.

\begin{enumerate} \itemsep4pt \parskip0pt \parsep0pt
\item Objective function $\bar{\phi}({\bar x}) :$ Problem dependent.

\item Metric $\bar{g}_{x}$:

The dominant computational cost comes from computing terms like $\mat{G}^T\mat{G}$, $\bar{\xi}_{\mat G}^T\bar{\eta}_{\mat G}$ and $\bar{\xi}_{\mat U}^T\bar{\eta}_{\mat U}$, each of these operations requires a numerical cost of $O(d_1 r^2)$. Other matrix operations  involve handling matrices of size $r \times r$ with total computational cost of $O(r^3)$.
\item Projecting on the tangent space $T_{\bar x} \overline{\mathcal W}$ with $\Psi_{\bar x} :$

It involves multiplications between matrices of sizes $d_1 \times r$ and $r \times r$ which costs $O(d_1 r^2)$. Other operations involve handling matrices of size $r \times r$.

\item Projecting on the horizontal space ${\mathcal H}_{\bar x}  \overline{\mathcal W}$ with $\Pi_{\bar x}$: 
\begin{itemize}
\item Forming the Lyapunov equations: Dominant computational cost of $O(d_1 r^2 + d_2 r^2)$ with matrix multiplications that cost $O(r^3)$. 
\item Solving the Lyapunov equations: $O(r^3)$ \cite{bartels72a}.
\end{itemize}
 
\item Retraction $R_{\bar x}$: 

\begin{itemize}
\item Computing the retraction on the $\Stiefel{r}{d_1}$ (the set of matrices of size $d_1 \times r$ with orthonormal columns) costs $O(d_1r^2)$
\item Computing the retraction on $\mathbb{R}_*^{d_1 \times r}$ costs $O(d_1 r)$

\item Computing the retraction on the set of positive-definite matrices $\ConePD{r}$ costs $O(r^3)$.
\end{itemize}

\item Riemannian gradient $\overline{\grad}_{\bar x} \bar{\phi}$:

First, it involves computing the partial derivatives of the cost function $\bar{\phi}$ which depend on the cost function $\bar{\phi}$. Second, the modifications to these partial derivatives involve matrix multiplications between matrices of sizes $d_1 \times r$ and $r \times r$ which costs $O(d_1 r^2)$.

\item Riemannian Hessian $\overline{\rc}_{\bar{\xi}_{\bar x}} \overline{ \grad _x \phi}$ in the direction $\bar{\xi}_{\bar x} \in \mathcal{H}_{\bar x} \overline{\mathcal{W}}$ on the total space:

The Riemannian Hessian on each of the three manifolds, $\Stiefel{d_1}{r}$, $\mathbb{R}_*^{d_1 \times r}$ and $\ConePD{r}$, consists of two terms. The first term is the Euclidean directional derivative of the Riemannian gradient in the direction $\bar{\xi}_{\bar x}$, i.e., $\D \overline{ \grad _x \phi} [\bar{\xi}_{\bar x}]$. The second term is the \emph{correction term} corresponds to the manifold structure and the metric. The summation of these terms is projected on the tangent space $T_{\bar x}\overline{\mathcal W}$ using $\Psi_{\bar x}$.
\begin{itemize}

\item{$\D \overline{ \grad _x \phi} [\bar{\xi}_{\bar x}]$: The computational cost depends on the cost function $\phi$ and its partial derivatives.}
\item{Correction term: It involves matrix multiplications with total cost of $O(d_1r^2 + r^3)$.}

\end{itemize}

\end{enumerate}
It is clear that all the geometry related operations are of linear complexity in $d_1$ and $d_2$; and cubic (or quadratic) in $r$. For the case of interest, $r \ll \min(d_1, d_2)$, these operations are therefore computationally very efficient. The ingredients that depend on the problem at hand are the evaluation of the cost function $\bar{\phi}$, computation of its partial derivatives and their directional derivatives along a search direction. In the next section, the computations of the partial derivatives and their directional derivatives are presented for the low-rank matrix completion problem.


\section{Numerical comparisons}\label{sec:numerical_comparisons}
In this section, we show numerical comparisons with the state-of-the-art algorithms. The application of choice is the low-rank matrix completion problem for which a number of algorithms with numerical codes are readily available. The competing algorithms are classified according to the way they view the set of fixed-rank matrices.

We show that our generic geometries connect closely with a number of competing methods. In addition to this, we bring out few conceptual differences between the competing algorithms and our geometric algorithms. Finally, the numerical comparisons suggest that our geometric algorithms compete favorably with the state-of-the-art.

\subsection{Matrix completion as a benchmark for numerical comparisons}\label{sec:matrix_completion}
To illustrate the notions presented in the paper, we consider the problem of low-rank matrix completion (described in Section \ref{sec:applic-completion}) as the benchmark application. The objective function is a smooth least square function and the search space is the space of fixed-rank matrices as shown in (\ref{eq:matrix-completion-formulation}). It is an optimization problem that has attracted a lot of attention in recent years. Consequently, a large body of algorithms have been proposed. Hence, this provides a good benchmark to not only compare different algorithms including our Riemannian geometric algorithms but also bring out the salient features of different algorithms and geometries. Rewriting the optimization formulation of the low-rank matrix completion, we have
\begin{equation}\label{eq:matrix_completion}
\begin{array}{llll}
	\min\limits_{ \mat{W}\in\FixedRank{r}{d_1}{d_2}}
		&	\frac{1}{|\Omega|}\|\mathcal{P}_{\Omega}(\mat{W}) - \mathcal{P}_{\Omega}(\mat{W}^{\star})\|_F^2 \\
\end{array}
\end{equation}  
where $\FixedRank{r}{d_1}{d_2}$ is the set of rank-$r$ matrices of size $d_1\times d_2$ and $\mat{W}^*$ is a matrix of size $d_1 \times d_2$ whose entries are given for indices $(i,j) \in \Omega$. $| \Omega |$ denotes the cardinality of the set $\Omega$ ($| \Omega | \ll  d_1 d_2$). $\mathcal{P}_{\Omega}$ is the orthogonal sampling operator, $\mathcal{P}_{\Omega}(\mat{W})_{ij}=\mat{W}_{ij}$ if $(i,j) \in \Omega$ and $\mathcal{P}_{\Omega}(\mat{W})_{ij}=0$ otherwise. We seek to learn a rank-$r$ matrix that best approximates the entries of $\mat{W}^*$ for the indices in $\Omega$.

As mentioned before, Table \ref{tab:spaces} to \ref{tab:gradient_Hessian} provide all the requisite information for implementing the (steepest) gradient descent and the Riemannian trust-region algorithms of Section \ref{sec:algorithms}. The only components still missing are the matrix formulae for the partial derivatives and their directional derivatives. These formulae are shown in Table \ref{tab:matrix_completion}. As regards the computational cost, the geometry related operations are linear in $d_2$ and $d_2$ (Section \ref{sec:numerical_complexity}); and the evaluation of the cost function, the computations of the partial derivatives and their directional derivatives depend \emph{primarily} on the computational cost of the auxiliary (sparse) variables $\mat{S}$ and $\mat{S}_*$ and the matrix multiplications of kind $\mat{SH}$ or $\mat{S}_* \mat{H}$ shown in Table \ref{tab:matrix_completion}. The variables $\mat{S}$ and $\mat{S}_*$ are respectively interpreted as the gradient of the cost function in the Euclidean space $\mathbb{R}^{d_1 \times d_2}$ and its directional derivative in the direction $\bar{\xi}_{\bar x}$. Finally, we have the following additional computation cost.
\begin{itemize}
\item Cost of computing $\bar{\phi}({\bar x})$: $O(|\Omega| r)$.
\item Computational cost of forming the sparse matrix $\mat{S}$: 

Computing the non-zero entries of $\mat{S}$ costs $O(|\Omega| r)$ plus the cost of updating of a sparse matrix for specific indices in $\Omega$. Both of these operations can be performed efficiently by MATLAB routines \cite{cai10a, wen10a, boumal11a}.

\item Computational cost of forming the sparse matrix $\mat{S}_*$: $O(|\Omega| r)$.
\item Computing the matrix multiplication $\mat{SH}$ or $\mat{S}_*\mat{H}$: 

Each costs $O(|\Omega|r)$. One gradient evaluation ($\grad_{\bar x}\bar{\phi}$) \emph{precisely} needs two such operations and a Hessian evaluation ($\overline{\rc}_{\bar{\xi}_{\bar x}} \overline{\grad_x \phi}$) needs four such operations.
\item Cost of computing all other matrix products: $O(d_1 r^2 + d_2 r^2 + r^3 )$.
\end{itemize}

\begin{table}
\begin{center} \scriptsize
\begin{tabular}{ p{1.0cm} | p{3.3cm} | p{4.0cm} | p{3.5cm} } 
& $\mat{W} = \mat{GH}^T$ & $\mat{W} = \mat{UBV}^T$ & $\mat{W} = \mat{UY}^T$ \\
\hline
& & & \\

Cost

 function $\bar{\phi}({\bar x})$
&  
$
\begin{array}[t]{lll}
\frac{1}{|\Omega|}\|\mathcal{P}_{\Omega}(\mat{GH}^T) \\
- \mathcal{P}_{\Omega}(\mat{W}^{\star})\|_F^2
\end{array}
$
 & 
$
\begin{array}[t]{lll}
\frac{1}{|\Omega|}\|\mathcal{P}_{\Omega}(\mat{UBV}^T) \\
- \mathcal{P}_{\Omega}(\mat{W}^{\star})\|_F^2
\end{array}
$
&
$ 
\begin{array}[t]{lll}
\frac{1}{|\Omega|}\|\mathcal{P}_{\Omega}(\mat{UY}^T) \\
- \mathcal{P}_{\Omega}(\mat{W}^{\star})\|_F^2
\end{array}
$
 \\ 
& & & \\

Partial 

derivatives of $\bar{\phi}$
&
$
\begin{array}[t]{lll}
(\mat{SH}, \mat{S}^T \mat{G}) \\
\in \mathbb{R}^{d_1 \times r} \times \mathbb{R}^{d_2 \times r}
\\
\end{array}
$

where

$
\begin{array}[t]{lll}
\\
\mat{S} = \frac{2}{|\Omega|}(\mathcal{P}_{\Omega}(\mat{GH}^T) \\
- \mathcal{P}_{\Omega}(\mat{W}^{\star}))
\end{array}
$
&
$
\begin{array}[t]{lll}
(\mat{SVB}, \mat{U}^T \mat{SV},\mat{S}^T \mat{UB}) \\
\in \mathbb{R}^{d_1 \times r} \times \mathbb{R}^{r \times r} \times \mathbb{R}^{d_2 \times r}
\\
\end{array}
$ 

where

$
\begin{array}[t]{lll}
\\
\mat{S} = \frac{2}{|\Omega|}(\mathcal{P}_{\Omega}(\mat{UBV}^T) \\
- \mathcal{P}_{\Omega}(\mat{W}^{\star}))
\end{array}
$
&
$
\begin{array}[t]{lll}
(\mat{SY}, \mat{S}^T \mat{U}) \\
\in \mathbb{R}^{d_1 \times r} \times \mathbb{R}^{d_2 \times r}

\end{array}
$

where 
$
\begin{array}[t]{lll}
\\
\mat{S} = \frac{2}{|\Omega|}(\mathcal{P}_{\Omega}(\mat{UY}^T) \\
- \mathcal{P}_{\Omega}(\mat{W}^{\star}))
\end{array}
$
\\

& & & \\

Riemannian 

gradient $\grad_{\bar x}\bar{\phi}$ from Table \ref{tab:gradient_Hessian} 
& 
$
\begin{array}[t]{lll}
(\mat{SH} \mat{G}^T \mat{G}, \mat{S}^T \mat{G}  \mat{H}^T \mat{H}) \\

\end{array}
$
& 
$
\begin{array}[t]{lll}
(\mat{SVB} - \mat{U}^T\Sym(\mat{U}^T \mat{SVB}), \\

\mat{B} \Sym(\mat{U}^T \mat{SV}) \mat{B}, \\

\mat{S}^T \mat{UB} - \mat{V}^T\Sym(\mat{V}^T \mat{S}^T \mat{UB} ) )
\end{array}
$ 
& 
$
\begin{array}[t]{lll}
(\mat{SY} - \mat{U}^T\Sym(\mat{U}^T \mat{SY}), \\

\mat{S}^T \mat{U} \mat{Y}^T \mat{Y})
\end{array}
$ 
\\
 & & & \\

 & & & \\
 
\hline 

\multicolumn{4}{c } { 

Directional derivative of the Riemannian gradient and its projection, i.e.,

$
\begin{array}{lll}
\\
\Psi_{\bar x}(\D \overline{ \grad _x \phi} [\bar{\xi}_{\bar x}]) \\
\\

\end{array}

$
}

\\

\hline
\\
\multicolumn{1}{ l | }{ $\mat{W} = \mat{GH}^T$  }
& 
\multicolumn{3}{l} {
$

\begin{array}[t]{lll}

 \Psi_{\bar x}( \mat{S}_*\mat{HG}^T\mat{G} + \mat{S}\bar{\xi}_{\mat H} \mat{G}^T \mat{G}  + 2\mat{SH}\Sym(\mat{G}^T \bar{\xi}_{\mat G}), \\
  \mat{S}_*^T \mat{GH}^T\mat{H} + \mat{S}^T \bar{\xi}_{\mat G} \mat{H}^T \mat{H}  + 2\mat{S}^T\mat{G}\Sym(\mat{H}^T \bar{\xi}_{\mat H})) \\
 
 \\
{\rm where} \ \mat{S}_* =  \frac{2}{|\Omega|}\mathcal{P}_{\Omega}(\mat{G}\bar{\xi}_{\mat H}^T +  \bar{\xi}_{\mat G} \mat{H}^T )

\end{array}
$

} \\

\\
\hline

\\
\multicolumn{1}{ l | }{ $\mat{W} = \mat{UBV}^T$  }
& 
\multicolumn{3}{l} {
$

\begin{array}[t]{lll}

 \Psi_{\bar x}( \mat{S}_* \mat{VB} + \mat{S}\bar{\xi}_{\mat V} \mat{B} + \mat{SV}\bar{\xi}_{\mat B} - \bar{\xi}_{\mat U} \Sym(\mat{U}^T \mat{SVB}), \\
 
 2\Sym(\mat{B}\Sym(\mat{U}^T \mat{SV}) \bar{\xi}_{\mat B} )  + \mat{B}\Sym(\bar{\xi}_{\mat U}^T \mat{SV} + \mat{U}^T \mat{S}_*\mat{V} + \mat{U}^T \mat{S} \bar{\xi}_{\mat V})\mat{B},\\
  
\mat{S}_*^T \mat{UB} + \mat{S}\bar{\xi}_{\mat U} \mat{B} + \mat{S}^T \mat{U} \bar{\xi}_{\mat B} - \bar{\xi}_{\mat V} \Sym(\mat{V}^T \mat{S}^T\mat{UB}))\\
 \\
{\rm where} \ \mat{S}_* =  \frac{2}{|\Omega|}\mathcal{P}_{\Omega}(     \mat{UB}\bar{\xi}_{\mat V}^T  + \mat{U} \bar{\xi}_{\mat B} \mat{V}^T  +   \bar{\xi}_{\mat U}\mat{BV}^T )

\end{array}
$

}
\\
 \\
\hline

\\
\multicolumn{1}{ l | }{ $\mat{W} = \mat{UY}^T$  }
& 
\multicolumn{3}{l} {
$

\begin{array}[t]{lll}
 \Psi_{\bar x}( \mat{S}_* \mat{Y} + \mat{S}\bar{\xi}_{\mat Y} - \bar{\xi}_{\mat U} \Sym(\mat{U}^T \mat{SY}), \\
  \mat{S}_*^T \mat{UY}^T\mat{Y} + \mat{S}^T \bar{\xi}_{\mat U} \mat{Y}^T \mat{Y}  + 2\mat{S}^T\mat{U}\Sym(\mat{Y}^T \bar{\xi}_{\mat Y})) \\
 \\
 
{\rm where} \ \mat{S}_* =  \frac{2}{|\Omega|}\mathcal{P}_{\Omega}(\mat{U}\bar{\xi}_{\mat Y}^T +  \bar{\xi}_{\mat U} \mat{Y}^T )

\end{array}
$

}
 \\
\hline
\end{tabular}
\end{center} 
\caption{Computation of the Riemannian gradient and its directional derivative in the direction $\bar{\xi}_{\bar x} \in \mathcal{H}_{\bar x} \overline{\mathcal W}$ for the low-rank matrix completion problem (\ref{eq:matrix_completion}). $\Psi_{\bar x}$ is the projection operator defined in Table \ref{tab:projections} and $\Sym(\cdot)$ extracts the symmetric part, $\Sym(\mat A) = \frac{\mat{A}^T + \mat{A}}{2}$. The development of these formulae follows systematically using the \emph{chain rule} of computing the derivatives. The auxiliary variables $\mat{S}$ and $\mat{S}_*$ are interpreted as the gradient of the cost function in the Euclidean space $\mathbb{R}^{d_1 \times d_2}$ and its directional derivative in the direction $\bar{\xi}_{\bar x}$ respectively.}
\label{tab:matrix_completion} 
\end{table}

All simulations are performed in MATLAB on a $2.53$ GHz Intel Core $\rm{i}5$ machine with $4$ GB of RAM. We use the MATLAB codes of all the competing algorithms supplied by their authors for our numerical studies. \change{For each example, a $d_1 \times d_2$ random matrix of rank $r$ is generated as in \cite{cai10a}. Two matrices $\mat{A} \in \mathbb{R}^{d_1 \times r}$ and $\mat{B} \in \mathbb{R}^{d_2 \times r}$ are generated according to a Gaussian distribution with zero mean and unit standard deviation. The matrix product $\mat{AB} ^T$ then gives a random matrix of rank $r$. A fraction of the entries are randomly removed with uniform probability}. \change{Note that the dimension of the space of $d_1 \times d_2$ matrices of rank $r$ is $(d_1 + d_2 - r)r$ and the number of known entries is a \emph{multiple} of this dimension. This multiple or ratio is called the \emph{over-sampling ratio} or simply, \emph{over-sampling} (OS)}. The over-sampling ratio (OS) determines the number of entries that are known. A $\rm{OS} = 6$ means that $6(d_1 + d_2 - r)r$ of randomly and uniformly selected entries are known a priori out of a total of $d_1d_2$ entries. \change{We use an initialization that is based on the rank-$r$ dominant singular value decomposition of $\mathcal{P}_{\Omega}(\mat{W}^*)$ \cite{boumal11a}. It should be stated that this procedure only provides a good initialization for the algorithms and we do not comment on the quality of this initialization procedure}. Numerical codes for the proposed algorithms for the low-rank matrix completion problem are available from the first author's homepage\footnote{\url{http://www.montefiore.ulg.ac.be/~mishra/pubs.html}.}. \change{Generic implementations of the three fixed-rank geometries can be found in the Manopt optimization toolbox \cite{boumal13a} which provides additional algorithmic implementations}.

All the considered gradient descent schemes, except RTRMC-$1$ \cite{boumal11a} and SVP \cite{jain10a}, use the adaptive step-size guess procedure (\ref{eq:adaptive_stepsize}) and the maximum number of iterations set at $200$. For the trust-region scheme, the maximum number of outer iterations is set at $100$ (we expect a better rate of convergence in terms of the outer iterations) and the number of inner iterations (for solving the trust-regions sub-problem) is bounded by $100$. Finally, the algorithms are stopped if the objective function value is below $10^{-20}$.

\change{In both the schemes we also set the initial step-size $s_0$ (for gradient descent) and the initial trust-region radius $\Delta _0$ (for trust-region) including the upper bound on the radius, $\bar{\Delta}$. We do this by \emph{linearizing} the search space. In particular, for the factorization $\mat{W} = \mat{UBV}^T$ (similarly for the other two factorizations) we solve the following optimization problem
\[
s_0 = \argmin\limits_{s} \|\mathcal{P}_{\Omega}( (\mat{U} - s \bar{\xi}_{\mat U} ) (\mat{B} - s\bar{\xi}_{\mat B} ) (\mat{V} - s \bar{\xi}_{\mat V} )^T) - \mathcal{P}_{\Omega}(\mat{W}^{\star})\|_F^2,  
\]
where $\bar{\xi}_{\bar x}$ is the Riemannian gradient. The above objective function is a degree $6$ polynomial in $s$ and thus, the global minimum $s_0$ can be obtained \emph{numerically} (and computationally efficiently) by finding the roots of a degree $5$ polynomial. $\Delta _0$ is then set to $\frac{s_0}{4^{3}}  \sqrt{\bar{g}_{\bar x}(\bar{\xi}_{\bar x}, \bar{\xi}_{\bar x})} $. The numerator of $\Delta  _0$ is the linearized trust-region radius and the reduction by $4^{3}$ considers the fact that this linearization might lead to an over-ambitious radius. Overall, this promotes a few extra gradient descent steps during the initial phase of the trust-region algorithm. The radii are upper-bounded as $\bar{\Delta} = 2^{10} \delta _0$. The integers $4$ and $2$ are used in the context of trust-region radius where an update is usually by a factor of $2$ and a reduction is by a factor of $4$ \cite[Algorithm~10]{absil08a}. The integers $3$ and $10$ have been chosen empirically.}

We consider the problem instance of completing a $32000 \times 32000$ matrix $\mat{W}^{\star}$ of rank $5$ as the running example in many comparisons. The over-sampling ratio OS is $8$ implying that  $0.25\%$ ($2.56 \times 10^{6}$ out of $1.04\times 10^{9}$) of entries are randomly and uniformly revealed. \change{In all the comparisons we show $5$ random instances to give a a more general comparative view. The over-sampling ratio of $8$ does not necessarily make the problem instance very challenging but it provides a standard benchmark to compare numerical scalability and performance of different algorithms. Similarly, a smaller tolerance is needed to observe the asymptotic rate of convergence of the algorithms.} \change{A rigorous comparison between different algorithms across different over sampling ratios and scenarios is beyond the scope of the present paper.}

\subsection{Full-rank factorization $\mat{W} = \mat{GH}^T$, MMMF, and LMaFit}\label{sec:balanced_update}
\begin{figure*}[t]
\subfigure[$\|\mat{H}_{0}\|_{F}\approx \|\mat{G}_{0}\|_{F}$.]{
\includegraphics[scale = .30]{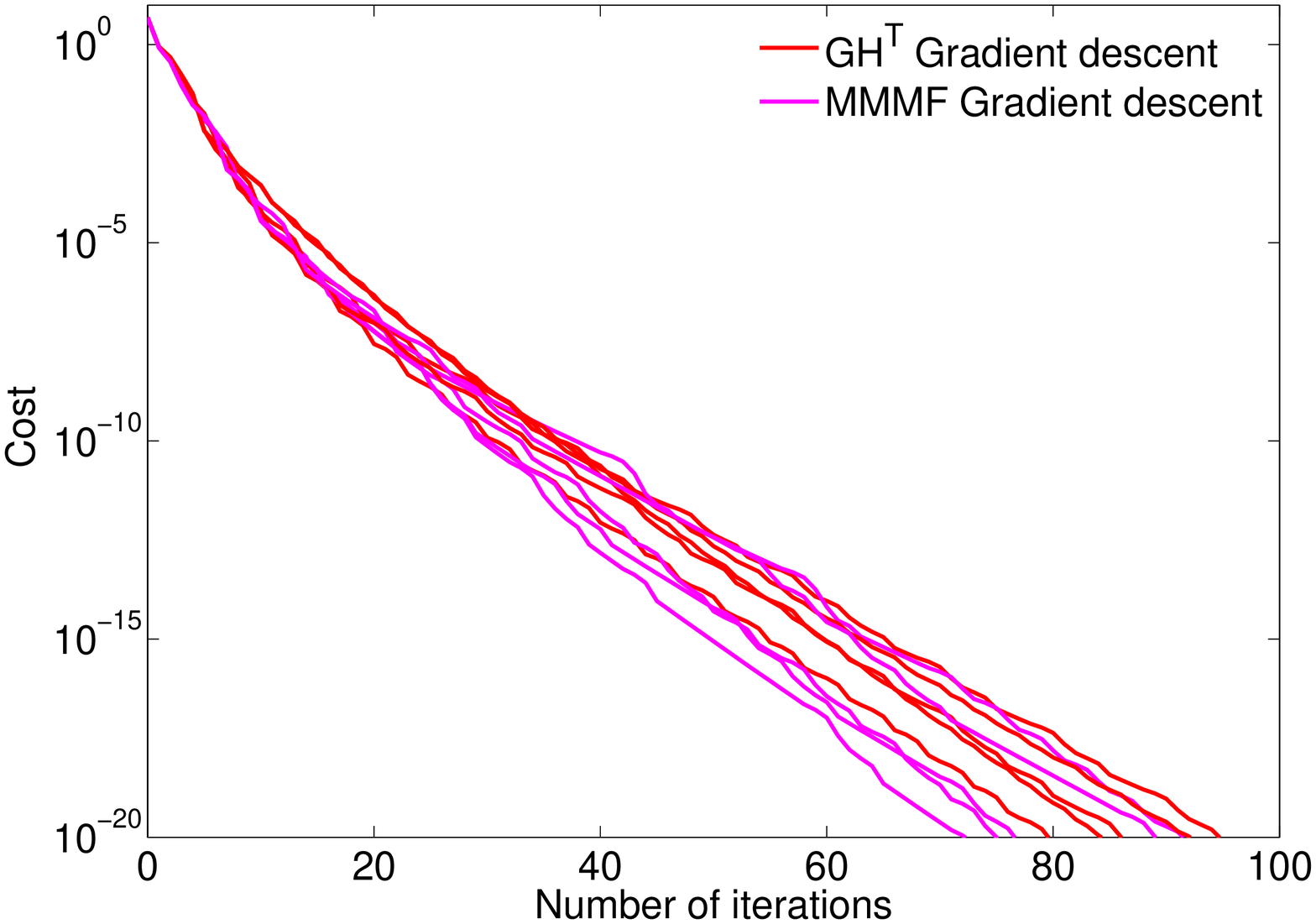}
}
\subfigure[$\|\mat{H}_{0}\|_{F}\approx 2 \|\mat{G}_{0}\|_{F}$.]{
\includegraphics[scale = .30]{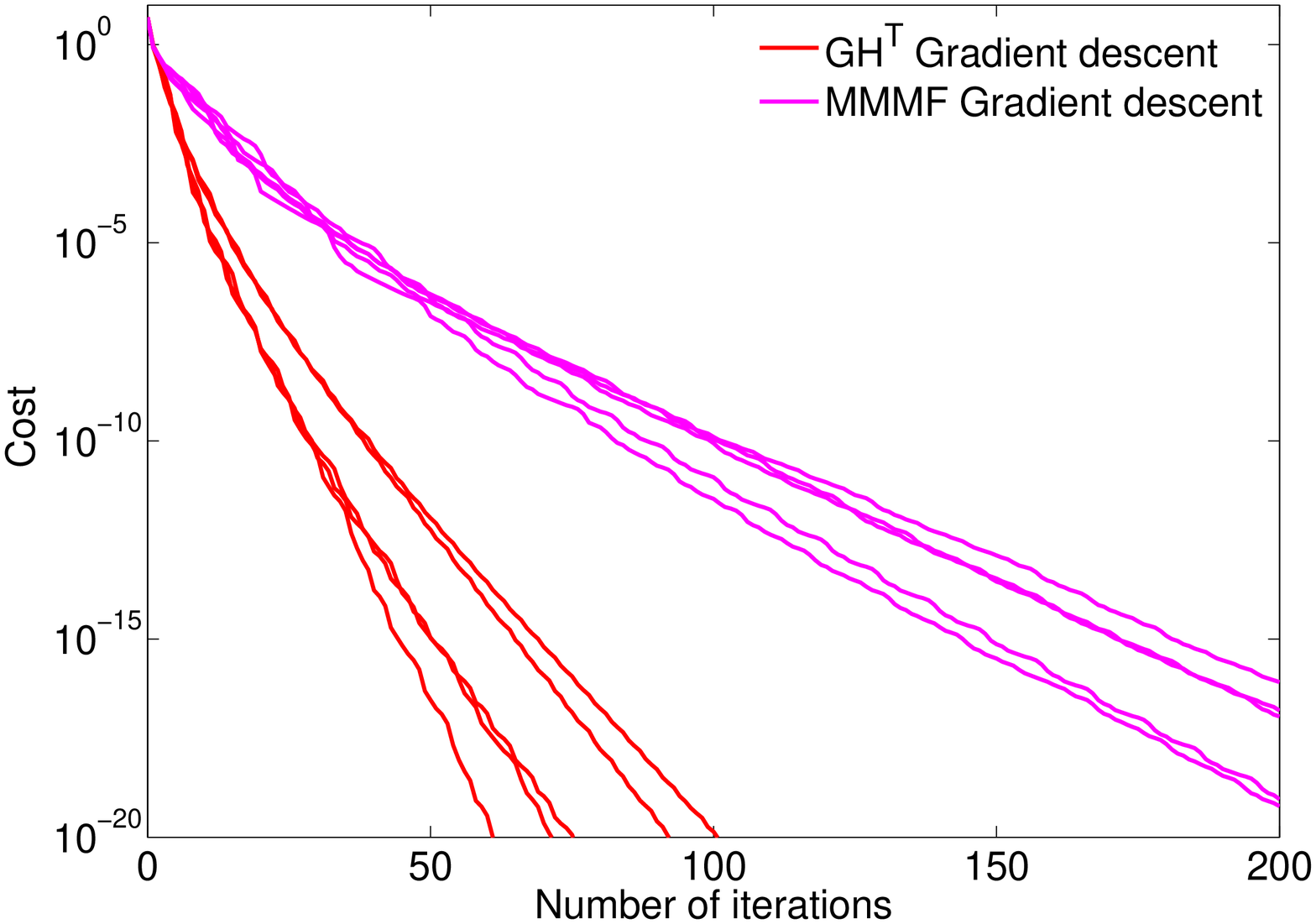}
}
\caption{$5$ random instances of low-rank matrix completion problems under different weights of factors at initialization. The proposed metric (\ref{eq:metric_gh}) resolves the issue of choosing an appropriate step-size when there is a discrepancy between  $\|\mat{G}\|_{F}$ and $\|\mat{H}\|_{F}$, a situation that leads to a slow convergence of the MMMF algorithm.}
\label{fig:step-issue}
\end{figure*}

The gradient descent algorithm for the full-rank factorization $\mat{W} = \mat{GH}^T$ is closely related to the gradient descent version of the Maximum Margin Matrix Factorization (MMMF) algorithm \cite{rennie05a}. The gradient descent version of MMMF is a descent step in the product space $\mathbb{R}_*^{d_1\times r} \times \mathbb{R}_*^{d_2 \times r}$ equipped with the Euclidean metric,
\begin{equation}\label{eq:metric_mmmf}
\begin{array}{lll}
\bar{g}_{\bar{x}}
 (    \bar{  \xi}_{\bar{x}} ,  \bar{\eta}_{\bar{x}}  )  & =  &\trace  (\bar{\xi}_{\mat G}^T  \bar{\eta}_{\mat G})   +  \trace (  \bar{\xi}^T_{\mat H} \bar{\eta}_{\mat H}  ) \\
\end{array}
\end{equation}
where $\bar{\xi}_{\bar{x}},\bar{\eta}_{\bar{x}} \in T_{\bar{x}} \overline{\mathcal{W}}$. Note the difference with respect to the metric proposed in Table \ref{tab:spaces} which is
\begin{equation}\label{eq:metric_gh}
\begin{array}{lll}
\bar{g}_{\bar{x}}
 (    \bar{  \xi}_{\bar{x}} ,  \bar{\eta}_{\bar{x}}  )  & =  &\trace  ((\mat{G}^T\mat{G})^{-1}\bar{\xi}_{\mat G}^T  \bar{\eta}_{\mat G})   +  \trace ( (\mat{H}^T\mat{H})^{-1} \bar{\xi}^T_{\mat H} \bar{\eta}_{\mat H}  ). \\
\end{array}
\end{equation}
As a result, the invariance (with respect to $r \times r$ non-singular matrices) is not taken into account in MMMF. In contrast, the proposed retraction in Table \ref{tab:projections} is invariant along the set of equivalence classes $\eqref{eq:equivalence-classes-balanced}$. This resolves the issue of choosing an appropriate step size when there is a discrepancy between $\|\mat{G}\|_{F}$ and $\|\mat{H}\|_{F}$. Indeed, this situation leads to a slower convergence of the MMMF algorithm, whereas the proposed algorithm is not affected (Figure \ref{fig:step-issue}). To illustrate this effect, we consider a rank $5$ matrix of size $4000  \times 4000$ with $2 \%$ of entries (${\rm OS} = 8$) are revealed uniformly at random. The Riemannian gradient descent algorithm based on the Riemannian metric (\ref{eq:metric_gh}) is compared against MMMF. In the first case, the factors at initialization has comparable weights, $\|\mat{H}_{0}\|_{F}\approx \|\mat{G}_{0}\|_{F}$. In the second case, we make factors at initialization slightly unbalanced, $\|\mat{H}_{0}\|_{F}\approx 2 \|\mat{G}_{0}\|_{F}$. \change{This discrepancy of the weights of the factors is not handled properly with the Euclidean metric (\ref{eq:metric_mmmf}) and hence, the rate of convergence of MMMF is affected as the plots show in Figure \ref{fig:step-issue}. The same also demonstrates that MMMF performs well when the factors are balanced. This understanding comes with notion of non-uniqueness of matrix factorization. In the previous example, though we force a bad balancing at initialization to show the relevance of scale-invariance, such a case might occur naturally for some particular cost functions and random initializations (e.g., when $d_2 \ll d_2 $). Hence, a discussion of choosing an appropriate metric has its merits.}

The LMaFit algorithm of \cite{wen10a} for the low-rank matrix completion problem also relies on the factorization $\mat{W} = \mat{GH}^T$ to alternatively learn the matrices $\mat{W}$, $\mat{G}$ and $\mat{H}$ so that the error $\| \mat{W}  - \mat{GH}^T \|^2_F$ is minimized while ensuring that the entries of $\mat{W}$ agree with the known entries, i.e., $\mathcal{P}_{\Omega}(\mat{W}) = \mathcal{P}_{\Omega}(\mat{W}^{\star})$. The algorithm is a tuned version the block-coordinate descent algorithm that has a \change{smaller} computational cost per iteration and better convergence than the standard non-linear \change{Gauss-Seidel} scheme.

\begin{figure*}[t]
\subfigure {
\includegraphics[scale = .30]{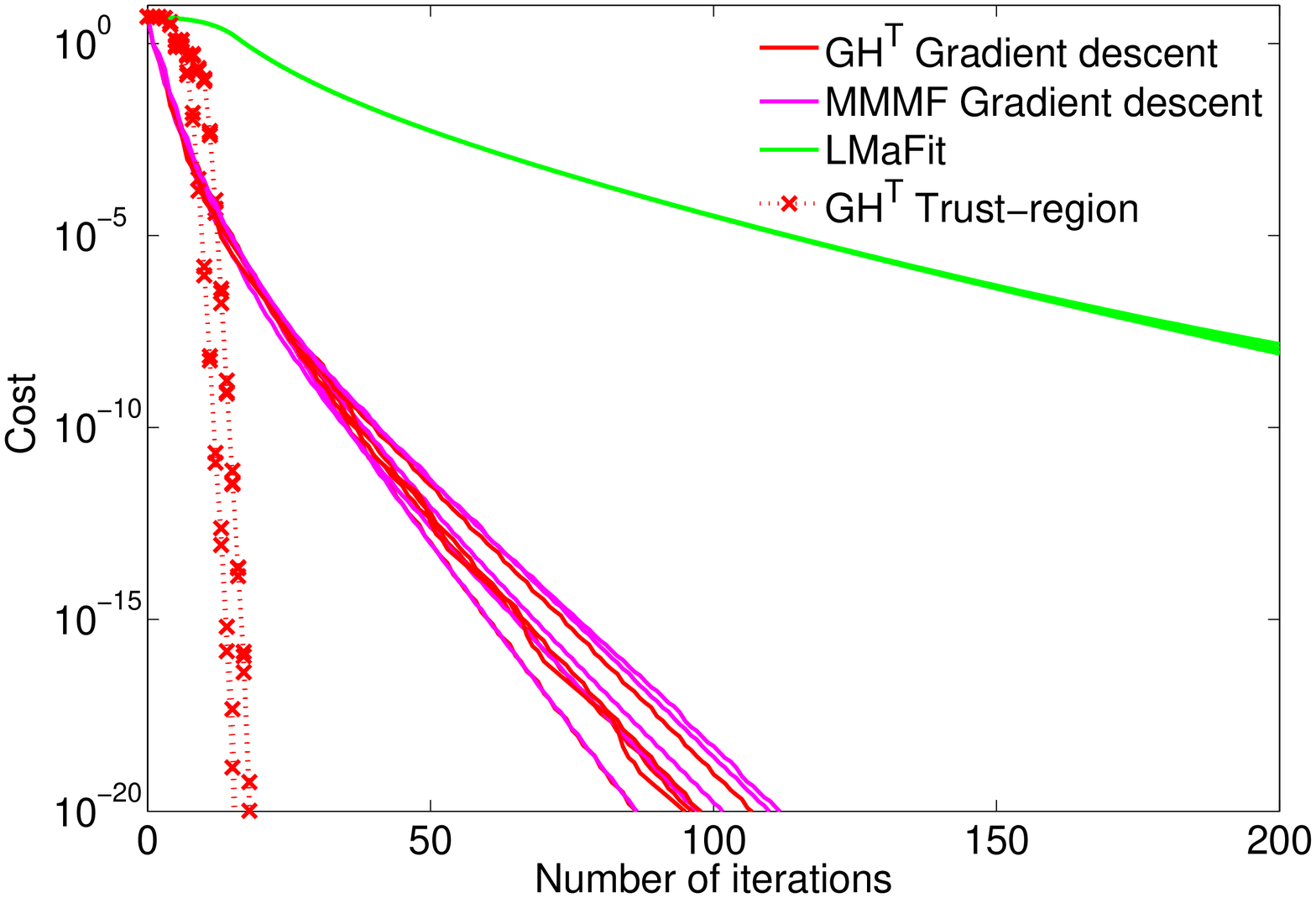}
}
\subfigure{
\includegraphics[scale = .30]{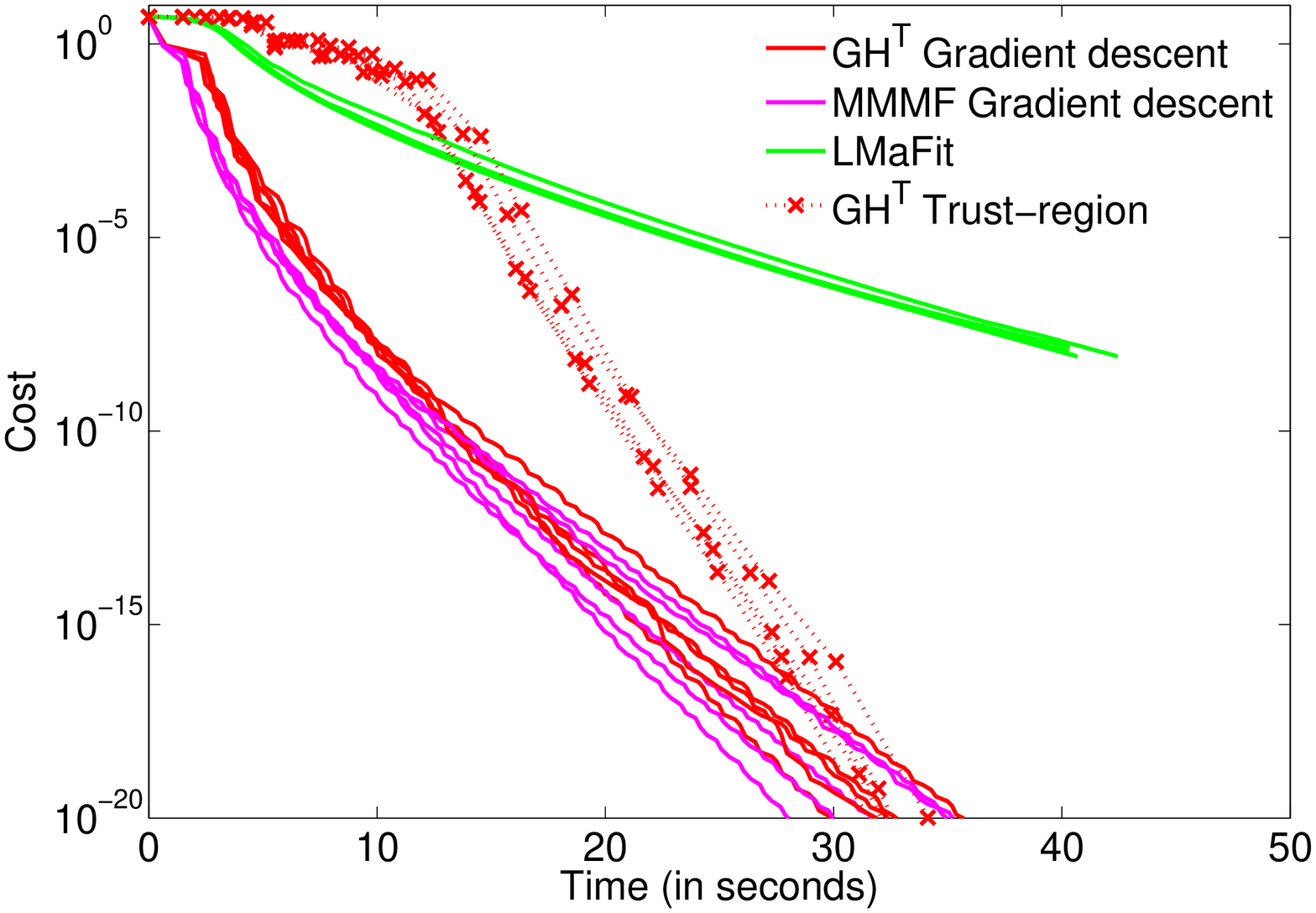}
}
\caption{$5$ random instances of rank $5$ completion of $32000 \times 32000$ matrix with ${\rm OS} = 8$. LMaFit has a smaller computational complexity per iteration but the convergence seems to suffer for large-scale matrices. MMMF and the gradient descent scheme perform similarly. After a slow start, the trust-region scheme shows a better rate of convergence.}
\label{fig:gd_vs_tr_gh}
\end{figure*}

We compare our Riemannian algorithms for the factorization $\mat{W} = \mat{GH}^T$ with LMaFit and MMMF in Figure \ref{fig:gd_vs_tr_gh}. Both MMMF and our gradient descent algorithm perform similarly. Asymptotically, the trust-region has a better rate of convergence both in terms of iterations and computational complexity. LMaFit reached $200$ iterations. During the initial few iterations, the trust-region algorithm adapts itself to the problem structure and takes non-effective steps where as the gradient descent algorithms are effective during the initial phase. Once in the region of convergence, the trust-region shows a better behavior. 

\subsection{Polar factorization $\mat{W} = \mat{UBV}^T$ and SVP}\label{sec:polar_vs_svd}
Here, we first illustrate the empirical evidence that constraining $\mat{B}$ to be diagonal (as is the case with singular value decomposition) is detrimental to optimization. We consider the simplest implementation of a gradient descent algorithm for matrix completion problem (see below). The plots shown in Figure \ref{fig:polar_vs_svd} compare the behavior of the same algorithm in the search space $\Stiefel{r}{d_1} \times \ConePD{r} \times \Stiefel{r}{d_2}$ (Section \ref{sec:factorizations-polar}) and $\Stiefel{r}{d_1} \times {\rm Diag}_{++}(r) \times \Stiefel{r}{d_2}$ (singular value decomposition). ${\rm Diag}_{++}(r)$ is the set of diagonal matrices of size $r \times r$ with positive entries. The metric and retraction updates are same for both the algorithms as shown in Table \ref{tab:projections}. The difference lies in constraining $\mat{B}$ to be diagonal which means that the Riemannian gradient for the later case is \change{also diagonal} and belongs to the space of $r\times r$ diagonal matrices, ${\rm Diag}(r)$ (the tangent space of the manifold ${\rm Diag}_{++}(r)$). \change{The matrix formulae for the factor $\mat{B}$ of the Riemannian gradient are therefore,
\[
\begin{array}{rll}
 \mat{B}\Sym(\mat{U}^T\mat{SV})\mat{B}& \ {\rm when\ } \mat{B} \in \ConePD{r}, \ {\rm and} \\
\mat{B}{\rm diag}(\mat{U}^T\mat{SV})\mat{B} &  \ {\rm when\ } \mat{B} \in  {\rm Diag}_{++}(r)
\end{array}
\]
where the notations are same as in Table \ref{tab:matrix_completion} and ${\rm diag}(\cdot)$ extracts the diagonal of a matrix, i.e., ${\rm diag}(\mat{A})$ is a diagonal matrix of size $r\times r$ with entries equal to the diagonal of $\mat{A}$.} The empirical observation that convergence suffers from imposing diagonalization on $\mat{B}$ is a generic observation and \change{has been noticed across various problem instances}. The problem here involves completing a $4000\times 4000$ of rank $5$ from $2\%$ of observed entries.
\begin{figure}[t]
\centering
\includegraphics[scale = 0.30]{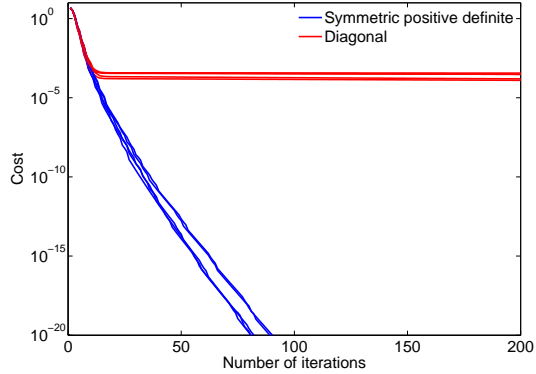}
\caption{Convergence of a gradient descent algorithm is affected by making $\mat{B}$ diagonal for the factorization $\mat{W} = \mat{UBV}^T$. The retraction updates for both the algorithms are same. The only difference is in the computation of the Riemannian gradient on the search space of ${\rm Diag}_{++}$ versus $\ConePD{r}$. The red curves reached $200$ iterations.}
\label{fig:polar_vs_svd}
\end{figure}

The OptSpace algorithm \cite{keshavan10a} also relies on the factorization $\mat{W}=\mat{U}\mat{B}\mat{V}^T$, but with $\mat{B}\in\mathbb{R}^{r\times r}$. At each iteration, the algorithm minimizes the cost function, say $\bar{\phi}$, by solving 
\begin{equation*}
	 \min_{\mat{U}, \mat{V}} \quad \bar{\phi}(\mat{U},\mat{B},\mat{V})
\end{equation*}
over the \change{\emph{bi}-Grassmann manifold}, $\Grassmann{r}{d_{1}} \times \Grassmann{r}{d_{2}}$ ($\Grassmann{r}{d_1}$ denotes the set of $r$-dimensional subspaces in $\mathbb{R}^{d_1}$) obtained by fixing $\mat{B}$ and then solving the inner optimization problem 
\begin{equation}\label{eq:inner_opt_optspace}
\min_{\mat{B}}\quad \bar{\phi}(\mat{U},\mat{B},\mat{V})
\end{equation}
for fixed $\mat{U}$ and $\mat{V}$. The algorithm thus alternates between a gradient descent step on the subspaces $\mat{U}$ and $\mat{V}$ for fixed $\mat{B}$, and a least-square estimation of $\mat{B}$ (matrix completion problem) for fixed $\mat{U}$ and $\mat{V}$. The proposed framework is different from OptSpace in the choice $\mat{B}$ positive definite versus $\mat{B}\in\mathbb{R}^{r\times r}$. As a consequence, each step of the algorithm retains the geometry of polar factorization. Our algorithm also differs from OptSpace in the simultaneous and progressive nature of the updates. A potential limitation of OptSpace comes from the fact that the inner optimization problem (\ref{eq:inner_opt_optspace}) may not be always solvable efficiently for other applications. 

The singular value projection (SVP) algorithm of \cite{jain10a} is based on the singular value decomposition (SVD) $\mat{W}=\mat{U}\mat{B}\mat{V}^T$ with $\mat{B}\in {\rm Diag}_{++}(r)$. It can also be interpreted in the considered framework as a gradient descent algorithm in the Euclidean space $\mathbb{R}^{d_1 \times d_2}$ (and hence, not the Riemannian gradient), along with an efficient SVD-projection based retraction exploiting the sparse structure of the gradient $\xi_{\rm Euclidean}$ (the gradient  in the Euclidean space $\mathbb{R}^{d_1 \times d_2}$, same as $\mat{S}$ in Table \ref{tab:matrix_completion}) for the matrix completion problem. A general update for SVP can be written as
\begin{equation*}
\mat{U}_{+} \mat{B}_{+}\mat{V}_{+}^{T}
= \SVD_{r}(\mat{U}\mat{B}\mat{V}^{T} - \xi_{\rm Euclidean}),
\end{equation*}
where $\SVD_{r}(\cdot)$ extracts the dominant $r$ singular values and singular vectors. An intrinsic limitation of the approach is that the computational cost of the algorithm is conditioned on the particular structure of the gradient. For instance, efficient routines exist for modifying the SVD with sparse \cite{larsen98a} or low-rank updates \cite{brand06a}. 

\begin{figure*}[t]
\subfigure {
\includegraphics[scale = .30]{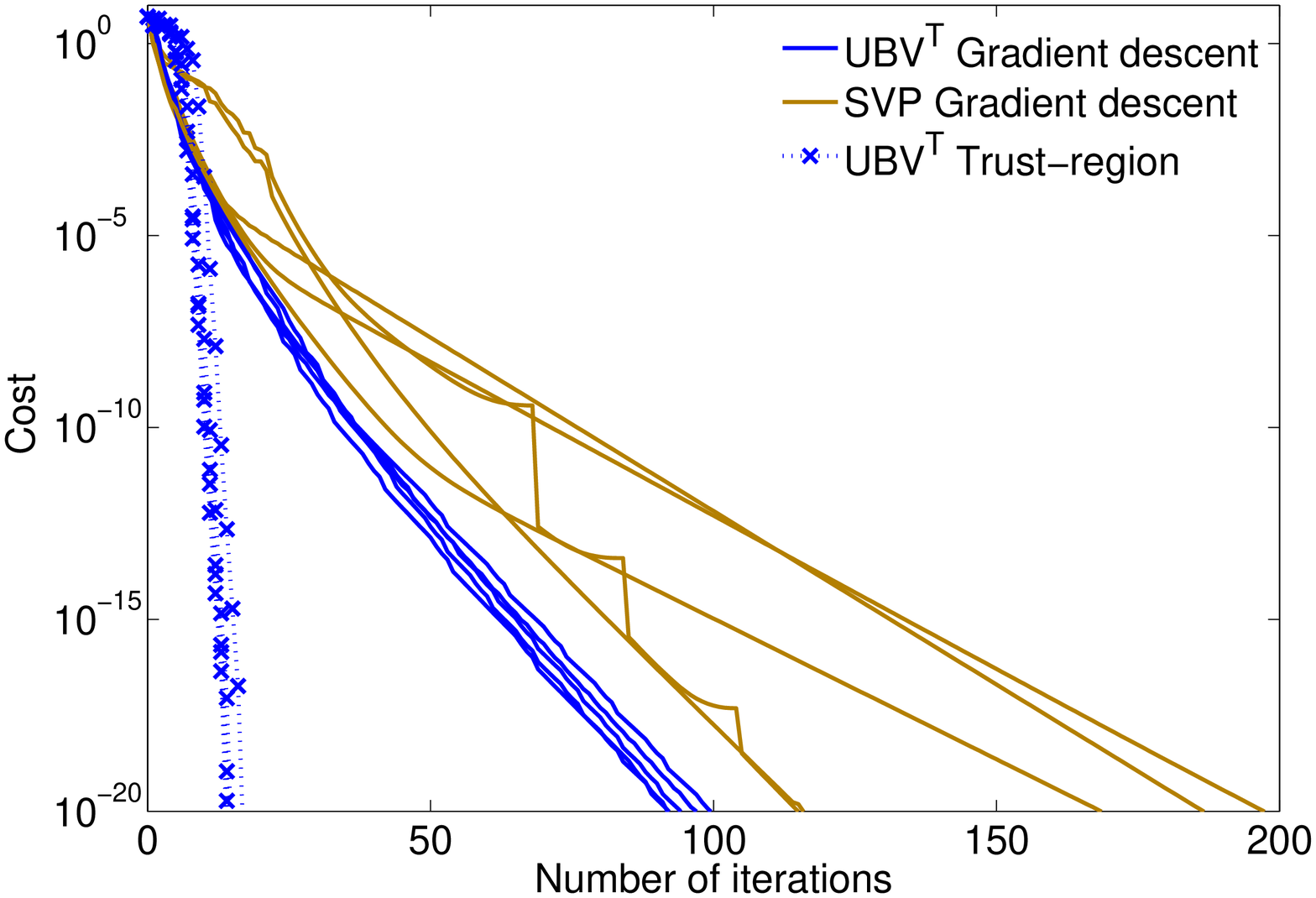}
}
\subfigure{
\includegraphics[scale = .30]{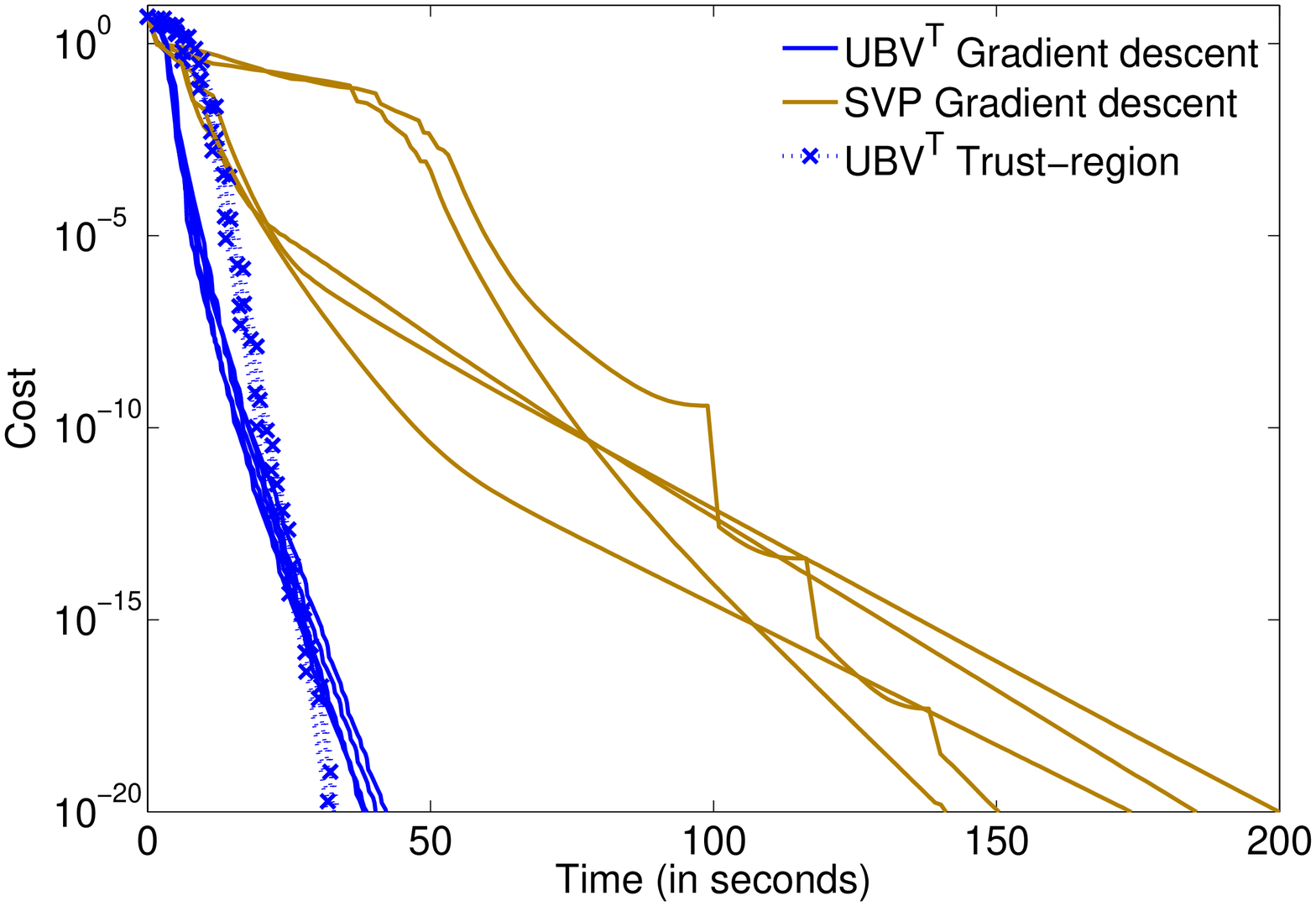}
}
\caption{Illustration of the Riemannian algorithms  on low-rank matrix completion problem for the factorization $\mat{W} = \mat{UBV}^T$ on $5$ random instances. Even though the number of iterations of SVP and our gradient descent are similar for some instances, the timings are very different. The main computational burden for SVP comes from computing the $r$ dominant singular value decomposition which is absent in the quotient geometry. Except for few sparse-matrix computations, most of our computations involve operations on dense matrices of sizes $d_1 \times r$ and $r \times r$ (Section \ref{sec:numerical_complexity}).}
\label{fig:ubv_svp}
\end{figure*}

\change{Both SVP and our gradient descent implementation use the Armijo backtracking method \cite[Procedure~3.1]{nocedal06a}. The difference is that for computing an initial step-size guess at each iteration SVP uses $\frac{d_1d_2}{ |\Omega|(1 + \delta)}$ with $\delta = 1/3$ as proposed in \cite{jain10a} while our gradient descent implementation uses the adaptive step-size procedure (\ref{eq:adaptive_stepsize})}. Figure \ref{fig:ubv_svp} shows the competitiveness of the proposed framework of factorization model $\mat{W} = \mat{UBV}^T$ with the SVP algorithm. Again, the trust-region asymptotically shows a better performance. The test example is an incomplete rank-$5$ matrix of size $32000\times 32000$ with ${\rm OS} = 8$. We could not compare the performance of the OptSpace algorithm as some MATLAB operations (in the code supplied by the authors) have not been optimized for large-scale matrices. We have, however, observed the good performance of the OptSpace algorithm on smaller size instances.

\subsection{Subspace-projection factorization $\mat{W} = \mat{UY}^T$ and RTRMC}\label{sec:invariance_metric_uy}
\begin{figure}[ht!]
	\centering
	\includegraphics[scale = .30]{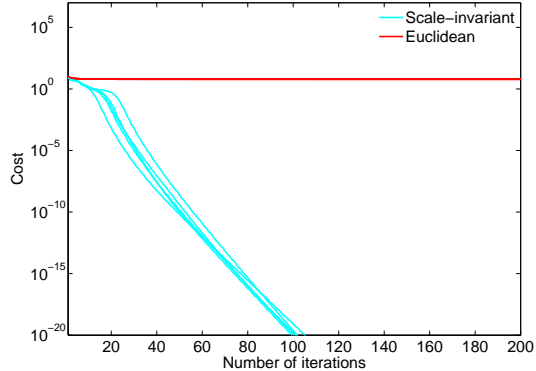}
	\caption{The choice of an scale-invariant metric (\ref{eq:metric_uy}) for subspace-projection factorization algorithm dramatically affects the performance of the algorithm. The algorithm with the Euclidean metric (\ref{eq:euclidean_uy}) flattens out due to a very slow rate of convergence because of numerical ill-conditioning due to the presence of unbalanced factors, $\|\mat{U}  \|_{F} \not \approx \|\mat{Y}  \|_{F}$. The example shown involves completing a rank-$5$ completion of a $4000\times 4000$ matrix with $98\%$ (${\rm OS}=8$) entries missing but the observation is generic.}
	\label{fig:affine_vs_flat}
\end{figure}

The choice of metric for the subspace-projection factorization shown in Table \ref{tab:spaces}, i.e.,
\begin{equation}\label{eq:metric_uy}
\begin{array}{lll}
\bar{g}_{\bar{x}}
 (     \bar{ \xi}_{\bar{x}} ,   \bar{\eta}_{\bar{x}}  )  & =  &\trace  (\bar{\xi}_{\mat U}^T  \bar{\eta}_{\mat U})   +  \trace ( (\mat{Y}^T\mat{Y})^{-1} \bar{\xi}^T_{\mat Y} \bar{\eta}_{\mat Y}  ) \\
\end{array}
\end{equation}
 is motivated by the fact that the total space $\Stiefel{r}{d_1} \times \mathbb{R}_* ^{d_2 \times r}$ equipped with the proposed metric is a complete Riemannian space and invariant to change of coordinates of the column space $\mat{Y}$. An alternative would be to consider the standard Euclidean metric for $\bar{\xi}_{\bar{x}},\bar{\eta}_{\bar{x}} \in T_{\bar{x}} \overline{\mathcal{W}}$,
\begin{equation}\label{eq:euclidean_uy}
\begin{array}{lll}
\bar{g}_{\bar{x}}
 (     \bar{ \xi}_{\bar{x}} ,   \bar{\eta}_{\bar{x}}  )  & =  &\trace  (\bar{\xi}_{\mat U}^T  \bar{\eta}_{\mat U})   +  \trace (  \bar{\xi}^T_{\mat Y} \bar{\eta}_{\mat Y}  ) \\
\end{array}
\end{equation}
which is also invariant by the group action $\OG{r}$ (the set of $r\times r$ matrices with orthonormal columns and rows) and thus, a valid Riemannian metric. This metric is for instance adopted in \cite{simonsson10a}, and \change{recently in \cite{absil12a} where the authors give a closed-form description of a \emph{purely} Riemannian Newton method}. Although this alternative choice is appealing for its numerical simplicity, Figure \ref{fig:affine_vs_flat} clearly illustrates the benefits of optimizing \change{with a metric that considers the scaling invariance property. The algorithm with the Euclidean metric (\ref{eq:euclidean_uy}) flattens out due to a very slow rate of convergence}. Under identical initializations and choice of step-size rule, our \change{proposed metric (\ref{eq:metric_uy})} prevents the numerical ill-conditioning \change{of the partial derivatives of the cost function (Table \ref{tab:matrix_completion})} that arises in the presence of unbalanced factors $\mat{U}$ and $\mat{Y}$, i.e., $\|\mat{U}  \|_{F} \not \approx \|\mat{Y}  \|_{F}$.  

The subspace-projection factorization is also exploited in the recent papers \cite{boumal11a, dai10a, dai10b} for the low-rank matrix completion problem. In the RTRMC algorithm of \cite{boumal11a} for the low-rank matrix completion problem the authors exploit the fact that in the variable $\mat{Y}$, $\min\limits_{\mat{Y}} \quad  \bar{\phi}(\mat{U}, \mat{Y})$ is a least square problem that has a closed-form solution. They are, thus, left with an optimization problem in the other variable $\mat{U}$ on the Grassmann manifold  $\Grassmann{r}{d_1}$.

\begin{figure*}[t]
\subfigure{
\includegraphics[scale = .30]{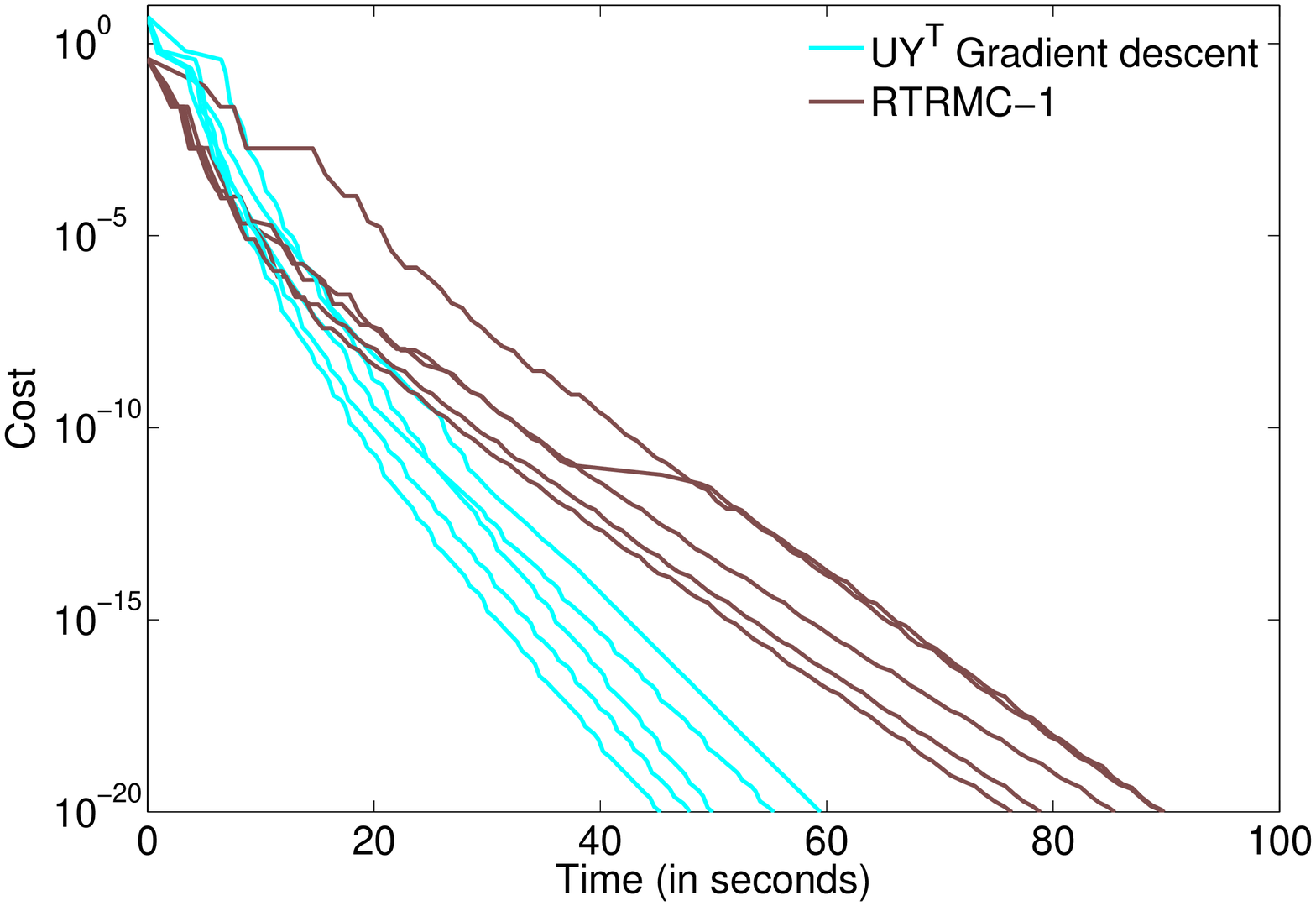}
}
\subfigure{
\includegraphics[scale = .30]{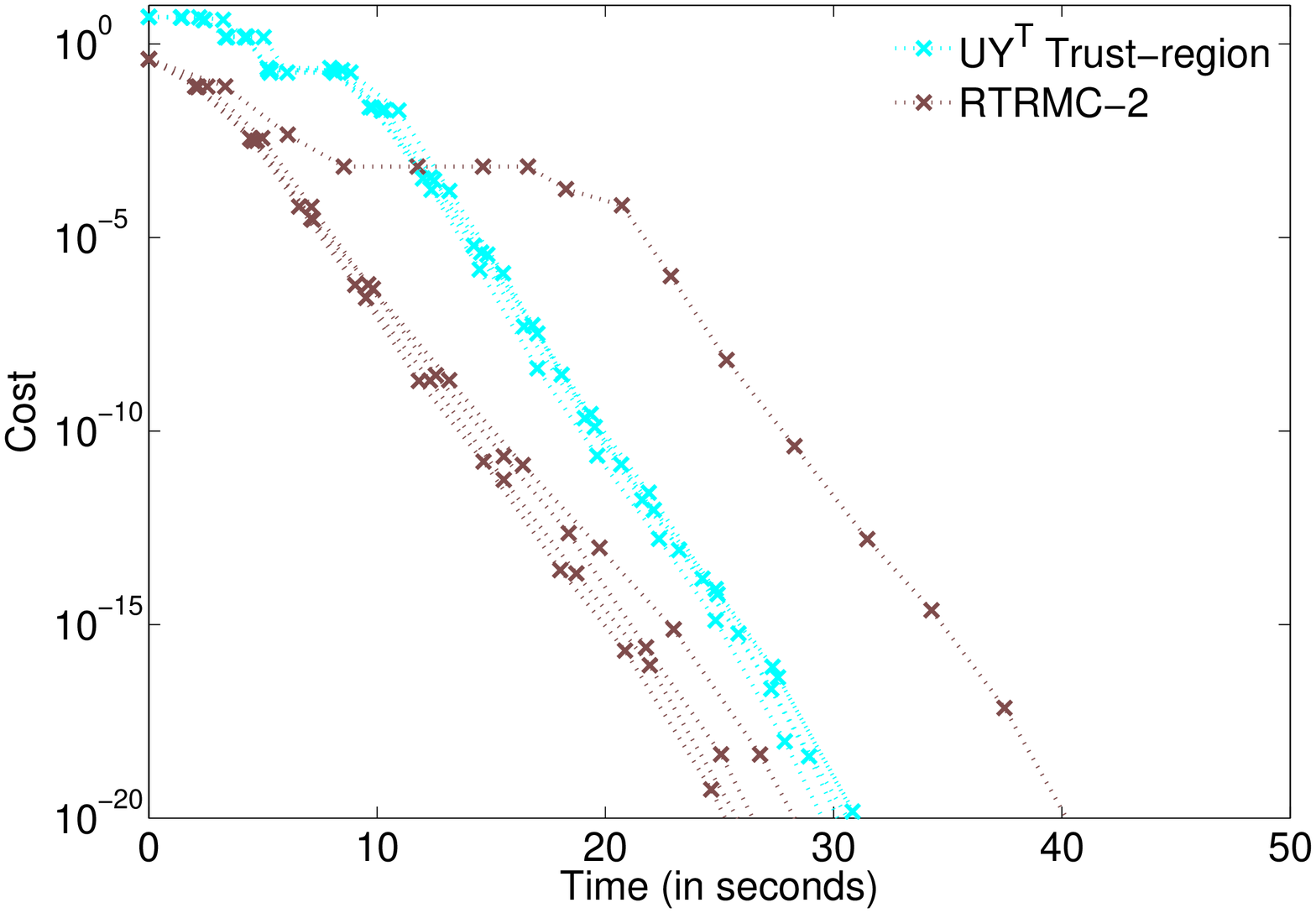}
}
\caption{$5$ random instances of rank-$5$ completion of $32000 \times 32000$ matrix with ${\rm OS} = 8$. The framework proposed in this paper is competitive with RTRMC when $d_1 \approx d_2$. For the trust-region algorithms, during the initial few iterations RTRMC-$2$ shows a better performance owing to the efficient least-square estimation of $\mat{Y}$. Asymptotically, both the algorithms perform similarly. For the gradient descent algorithms, however, our implementation shows a better timing performance.}
\label{fig:uz_vs_rtrmc}
\end{figure*}

The resulting geometry of RTRMC is efficient in situations where $d_1 \ll d_2$ where the least square is solved efficiently in the dimension $d_2  r$ and the optimization problem is on a smaller search space of dimension $d_1 r - r^2$. The advantage is reduced in square problems and the numerical experiments in Figure \ref{fig:uz_vs_rtrmc} suggest that our generic algorithm compares favorably to the Grassmanian algorithm in \cite{boumal11a} in that case. \change{Similar to our trust-region algorithm, RTRMC-$2$ is a trust-region implementation with the parameters $\theta = 1$ and $\kappa = 0.1$ (Section \ref{sec:trust_region}). The parameters $\Delta _0$ and $\bar{\Delta}$ are chosen as suggested in \cite{boumal12a}. Both RTRMC and our trust-region algorithm use the solver GenRTR \cite{genrtr} to solve the trust-region sub-problem. RTRMC-$1$ is RTRMC-$2$ with the Hessian replaced by identity that yields the steepest descent algorithm. The number of iterations needed by both the algorithms are similar and hence, not shown in Figure \ref{fig:uz_vs_rtrmc}.}

\subsection{Quotient and embedded viewpoints}\label{sec:quotient_embedded}
\change{In Section \ref{sec:quotient_spaces} we have viewed the set of fixed-rank matrices as the product space of well-studied manifolds $\Stiefel{r}{d_1}$ (the set of matrices of size $d_1 \times r$ with orthonormal columns), $\mathbb{R}_*^{d_1 \times r}$ (the set of matrices of size $d_1 \times r$ with full column rank) and $\ConePD{r}$ (the set of positive definite matrices of size $r\times r$) and consequently, the search space admitted a Riemannian quotient manifold structure.}
A different viewpoint is that of the \emph{embedded submanifold} approach. \change{The search space $\FixedRank{r}{d_1}{d_2}$ (the set of rank-$r$ matrices of size ${d_1 \times d_2})$ admits a Riemannian submanifold of the Euclidean space $\mathbb{R}^{d_1 \times d_2}$} \cite[Proposition~2.1]{vandereycken13a}. Recent papers \cite{vandereycken13a, shalit10a} investigate the search space in detail and develop the notions of optimizing a smooth cost function. While conceptually the iterates move on the embedded submanifold, numerically the implementation is done using factorization models, \change{the full-rank factorization is used in \cite{shalit10a} and a compact singular value decomposition is used in \cite{vandereycken13a}.}

\begin{table}
\begin{center} \scriptsize
\begin{tabular}{ p{4cm} | p{8cm} } 
& Embedded submanifold $\FixedRank{r}{d_1}{d_2}$  \\
\hline

  &  \\

 Matrix representation & 
$\mat{W} = \mat{U\Sigma} \mat{V}^T $

where $\mat{U} \in \Stiefel{r}{d_1}$, $\mat{\Sigma} \in {\rm Diag}_{++}(r)$, and $\mat{V}\in \Stiefel{r}{d_2}$
  \\

  &  \\

Tangent space 

$T_{\mat W} \FixedRank{r}{d_1}{d_2}$
& 
$
\begin{array}[t]{lll}

	\mat{U}\mat{N}\mat{V}^{T} + \mat{U}_p\mat{V}^{T} + \mat{U} \mat{V}_p ^{T} :  \mat{N}\in\mathbb{R}^{r\times r}, \\
	\mat{U}_p \in\mathbb{R}^{d_1 \times r}, \mat{U}_{p}^T \mat{U} = \mat{0} , \\ 
	 \mat{V}_p \in\mathbb{R}^{d_2 \times r}, \mat{V}_{p}^T \mat{V} = \mat{0} \\
	
\end{array}
$

\\
  &  \\

Metric $g_{\mat W}(\mat{Z}_1, \mat{Z}_2)$&  
$
\begin{array}[t]{ll}
\trace(\mat{Z}_1^T \mat{Z}_2)
\end{array}
$

   \\
   
   &  \\

Projection of a matrix $\mat{Z} \in \mathbb{R}^{d_1 \times d_2}$ onto the tangent space $T_{\mat W} \FixedRank{r}{d_1}{d_2}$ & 
$
\begin{array}[t]{lll}
\Pi_{\mat W} (\mat{Z}) = \{ 
\mat{P}_{\mat {U}}  \mat{Z} \mat{ P}_{\mat{V}} + \mat{P}_{\mat{U}}^{\perp} \mat{Z} \mat{P}_{\mat V}  + \mat{P}_{\mat{U}} \mat{Z} \mat{P}_{\mat V} ^{\perp} :\\
\mat{P}_{\mat{U}} := \mat{UU}^T$ and $\mat{P}_{\mat{U}}  ^{\perp}:= \mat{I} - \mat{P}_{\mat U} \}
\end{array}
$

\\
  &  \\

Riemannian gradient &

$\grad_{\mat W} f = \Pi_{\mat{W}} (\Grad_{\mat{W}} \bar{f} )$

where $\Grad_{\mat W} \bar{f}$ is the gradient of $\bar f$ in $\mathbb{R}^{d_1 \times d_2}$

 \\

 & \\

 Riemannian connection 
 
 $ {\rc}_{\xi} \eta $ where 
 
 $\xi , \eta \in T_{\mat W} \FixedRank{r}{d_1}{d_2}$  \cite[Proposition~5.3.2]{absil08a}
 
 &
$
\Pi_{\mat W} ( \D \bar{\eta}  [\bar{\xi}] )
$

where,
$\D \bar{\eta}  [\bar{\xi}]$ is the standard Euclidean directional derivative of $\bar{\eta}$ in the direction $\bar{\xi}$
  
  \\

  &  \\

Retraction  & 
$
R_{\mat W} (\xi) =  \rm{SVD} (\mat{W} + \xi)
$
where ${\rm SVD}$ involves the computation of a thin singular value decomposition
with rank $2r$ \\  
  
  &  \\
  
  \hline

\end{tabular}
\end{center} 
\caption{Optimization-related ingredients for using the embedded geometry of rank-$r$ matrices at $\mat{W} \in \FixedRank{r}{d_1}{d_2}$ \cite{vandereycken13a}. The rank-$r$ matrix $\mat{W}$ is stored in the factorized form $(\mat{U}, \mat{\Sigma}, \mat{V})$ resulting form a compact singular value decomposition. As a consequence, it leads to computationally efficient calculations of all the above listed ingredients. The computation of the Riemannian gradient is shown for a smooth cost function $\bar{f}: \mathbb{R}^{d_1 \times d_2} \rightarrow \mathbb{R}$ and its restriction $f$ on the manifold $\FixedRank{r}{d_1}{d_2}$. 
} 
\label{tab:ingredients_embedded} 
\end{table} 

The characterization of the embedded geometry is tabulated in Table \ref{tab:ingredients_embedded} using the factorization model $\mat{W}=\mat{U} \mat{\Sigma} \mat{V}^{T}$. Here $\mat{\Sigma} \in {\rm Diag}_{++}$ is a diagonal matrix with positive entries, $\mat{U} \in \Stiefel{r}{d_1}$ and $\mat{V} \in \Stiefel{r}{d_2}$. The treatment is similar for the factorization $\mat{W} = \mat{GH}^T$ as the underlying geometries are same \cite{shalit10a}.

The visualization of the search space as an embedded submanifold of $\mathbb{R}^{d_1 \times d_2}$ has some key advantages. For example, the notions of geometric objects can be interpreted in a straight forward way. In the matrix completion problem, this also allows us to compute the initial step-size guess (in a search direction) by linearizing the search space \cite{vandereycken13a}. On the other hand, the product space representation of Section \ref{sec:quotient_spaces} of fixed-rank matrices seems naturally related to matrix factorization and provides additional flexibility in choosing the metric. It is only the horizontal space (Section \ref{sec:horizontal_lifts}) that couples the product spaces. From the optimization point of view this flexibility is also of interest. For instance, it allows us to regularize the matrix factors, say $\mat{G}$ and $\mat{H}$, differently.

\begin{figure*}[t]
\subfigure{
\includegraphics[scale = .30]{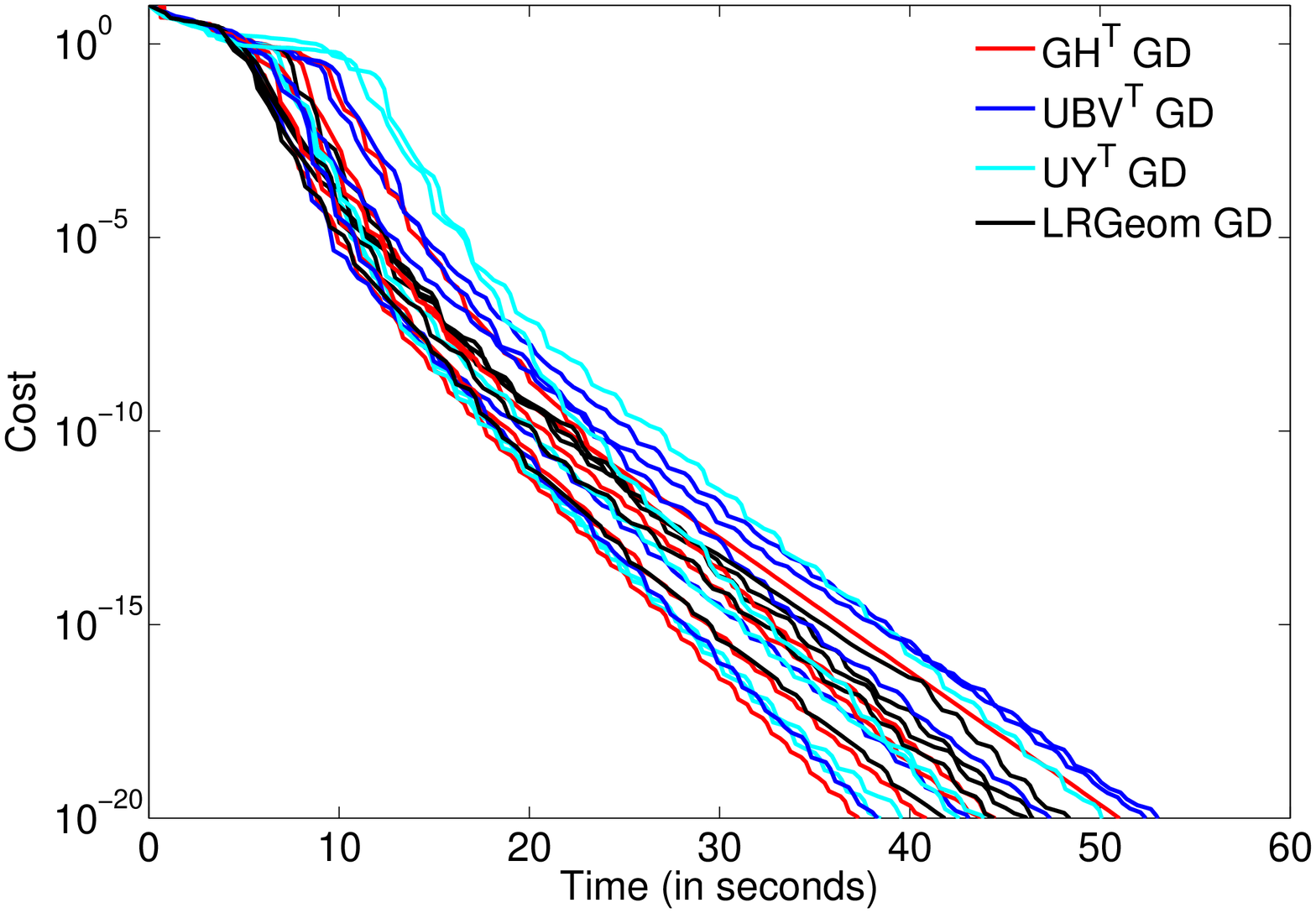}
}
\subfigure{
\includegraphics[scale = .30]{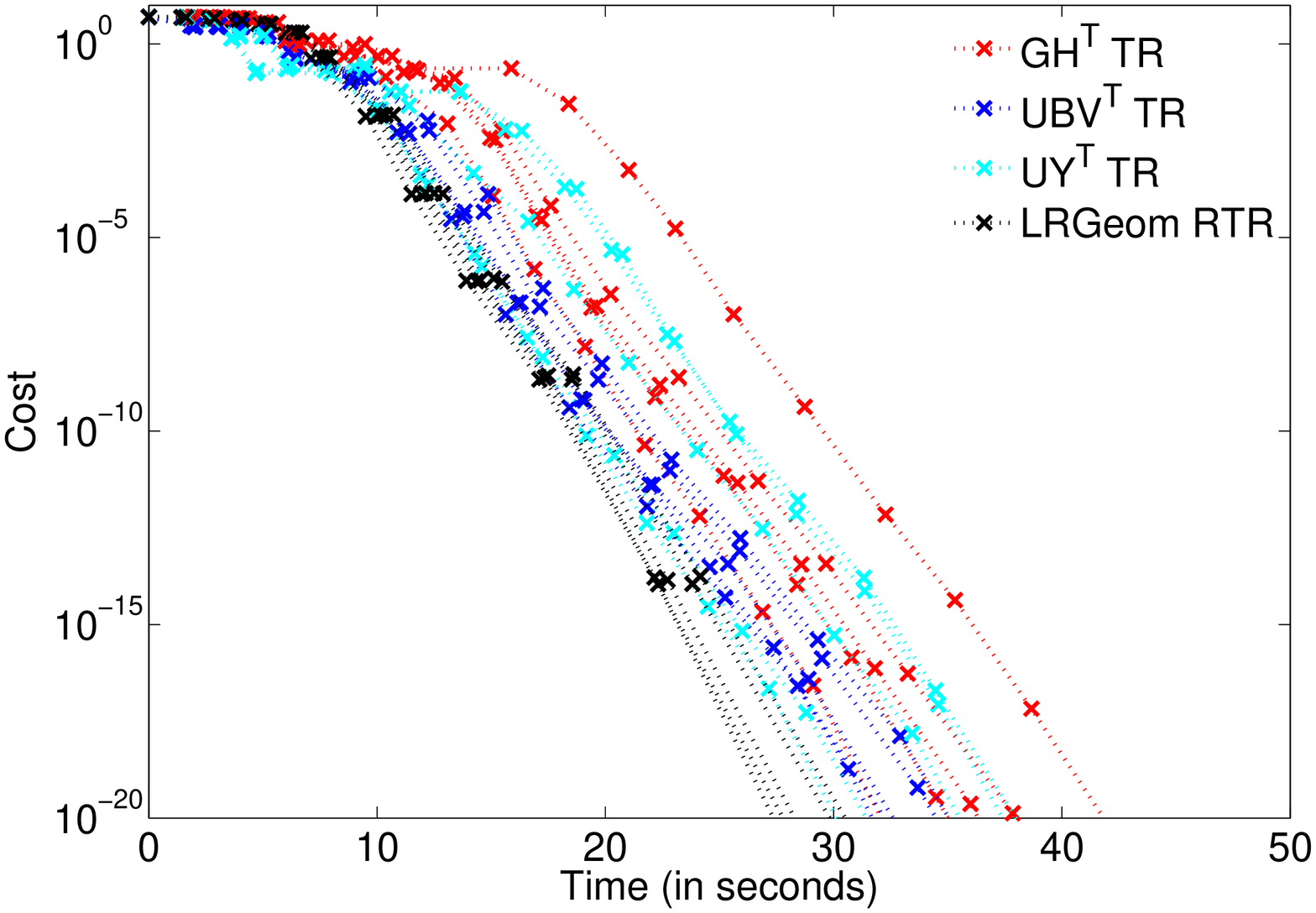}
}
\caption{Low-rank matrix completion of size $32000\times 32000$ of rank $5$ with ${\rm OS} = 8$. Both quotient and embedded geometries behave similarly. The behaviors of the gradient descent (GD) algorithms of these geometries are indistinguishable. The trust-region (TR) schemes perform similarly with LRGeom RTR showing a better performance during the initial few iterations.}
\label{fig:quotient_vs_embedded}
\end{figure*}

\change{In Figure \ref{fig:quotient_vs_embedded} we compare our algorithms with LRGeom (the algorithmic implementation of \cite{vandereycken13a}) on $5$ random instances.  The timing plots for gradient descent and trust-region algorithms show that Riemannian quotient algorithms are competitive with LRGeom. The parameters $s_0$, $\Delta_0$ and $\bar{\Delta}$ for all the algorithms are set by performing a linearized search as proposed in Section \ref{sec:matrix_completion}. The linearized step-size search for LRGeom is the one proposed in \cite{vandereycken13a}. LRGeom RTR (the trust-region implementation) shows a better performance during the initial phase of the algorithm. The trust-region schemes based on the quotient geometries seem to spend more time \emph{in transition} to the region of rapid convergence. However asymptotically, we obtain the same performance as that of LRGeom RTR. The behaviors of all the gradient descent schemes are inseparable.}

\section{Conclusion}
We have addressed the problem of rank-constrained optimization (\ref{eq:intro-general-formulation}) and presented both first-order and second-order schemes. The proposed framework is general and encompasses recent advances in optimization algorithms. We have shown that classical fixed-rank matrix factorizations have a natural interpretation of classes of equivalences in well-studied manifolds. As a consequence, they lead to a matrix search space that has the geometric structure of a Riemannian submersion, with convenient matrix expressions for all the geometric objects required for an optimization algorithm.   The computational cost of involved matrix operations is always linear in the original dimensions of the matrix, which makes the proposed computational framework amenable to large-scale applications. The product structure of the considered total spaces provides some flexibility in choosing the proper metrics  on the search space. The relevance of this flexibility was illustrated    in the context of subspace-projection factorization $\mat{W} = \mat{UY}^T$ in Section \ref{sec:invariance_metric_uy}. The relevance of not fixing the matrix factorization beyond necessity has been illustrated  in the context of $\mat{W} = \mat{UBV}^T$ factorization in Section \ref{sec:polar_vs_svd} where the flexibility of $\mat{B}$ to be positive definite instead of diagonal (as is the case with singular value decomposition) results in good convergence properties. Similarly, the advantage of balancing an update for $\mat{W} = \mat{GH}^T$ factorization has been discussed in Section \ref{sec:balanced_update}. 

All numerical illustrations of the paper were provided on the low-rank matrix completion problem, that permitted a comparison with many existing fixed-rank optimization algorithms. It was shown that the proposed framework  compares favorably  with most state-of-the-art algorithms  while maintaining a complete generality.

\change{The three considered geometries show a comparable numerical performance in the simple examples considered in the paper. However, differences exist in the resulting metrics and related invariance properties, which may lead to a geometry being preferred for a particular problem. In the same way as different matrix factorizations exist and the preference for one factorization is problem dependent, we view the three proposed geometries as three possible choices which the user should exploit as a source of flexibility in the design of a particular optimization algorithm tuned to a particular problem. They are all equivalent in terms of numerical complexity and convergence guarantees.}

Optimizing the geometry and the metric to a particular problem such as matrix completion and to a particular dataset will be the topic of future research. Some  steps in that direction are proposed in the recent papers \cite{ngo12a, mishra12a}.

\bibliographystyle{amsalpha}
\bibliography{arXiv_MMBS_CoST_2012}



\end{document}